\documentclass[a4paper,review]{cas-sc}
\usepackage[numbers]{natbib}

\usepackage{amssymb}
\usepackage{times}
\usepackage{epsfig}
\usepackage{graphicx}
\usepackage{amsmath}

\usepackage{amsthm}
\usepackage{booktabs}
\usepackage{algorithm}
\usepackage{algorithmic}
\usepackage{xcolor}
\usepackage{multirow}
\usepackage{subcaption}

\usepackage[justification=centering]{caption}
\usepackage{makecell}
\usepackage{utfsym}
\usepackage{colortbl}
\usepackage{amsfonts} 
\usepackage[section]{placeins}
\usepackage{microtype}
\usepackage[figuresright]{rotating}
\usepackage{setspace}

\newtheorem{proposition}{Proposition}

\begin{document}

\let\WriteBookmarks\relax
\def\floatpagepagefraction{1}
\def\textpagefraction{.001}
\shorttitle{Contrast and Clustering} 
\title [mode = title]{Contrast and Clustering: Learning Neighborhood Pair Representation for Source-free Domain Adaptation}            
\shortauthors{Yuqi Chen}  

\author[jxnu,zjnu]{Yuqi Chen}[orcid=0000-0001-9769-1167]
\ead{202025201103@zjnu.edu.cn}

\author[zjnu]{Xiangbin Zhu}[orcid=0000-0001-7281-6680]
\ead{zhuxb@zjnu.cn}

\author[jxnu]{Yonggang Li}[orcid=0000-0002-7269-0368]
\ead{liyonggang@zjxu.edu.cn}
\cormark[1]

\author[zjnu]{Yingjian Li}
\ead{liyingjian@zjnu.edu.cn}
\author[zlg]{Haojie Fang}
\ead{18868901971@163.com}

\address[jxnu]{College of Information Science and Engineering, Jiaxing University, Jiaxing, 314001, China}
\address[zjnu]{School of Computer Science and Technology, Zhejiang Normal University, Jinhua, 321004, China}

\address[zlg]{School of Computer Science and Technology, Zhejiang Sci-Tech University, Hangzhou, 310018, China}

\begin{abstract}
Unsupervised domain adaptation aims to address the challenge of classifying data from unlabeled target domains by leveraging source data from different distributions. However, conventional methods often necessitate access to source data, raising concerns about data privacy. In this paper, we tackle a more practical yet challenging scenario where the source domain data is unavailable, and the target domain data remains unlabeled. To address the domain discrepancy problem, we propose a novel approach from the perspective of contrastive learning. Our key idea revolves around learning a domain-invariant feature by: 1) constructing abundant pairs for feature learning by utilizing neighboring samples. 2) Refining negative pairs pool reduces learning confusion; and 3) combining noise-contrastive theory simplifies the function effectively. Through careful ablation studies and extensive experiments on three common benchmarks, VisDA, Office-Home, and Office-31, we demonstrate the superiority of our method over other state-of-the-art works. Our proposed approach not only offers practicality by alleviating the requirement of source domain data but also achieves remarkable performance in handling domain adaptation challenges. The code is available at https://github.com/yukilulu/CaC.
\end{abstract}




\maketitle


\section{Introduction}
The demand for extensive labeled training data has been effectively met by unsupervised learning. Nevertheless, a considerable decmidrule in performance can be observed when the data distributions in the source and target domains exhibit substantial differences, formally known as domain or distribution shift. To tackle this issue, domain adaptation (DA) methods \cite{caco,da:adv} rely on the co-training of source and target data, aiming to transfer learned knowledge from the source domain to the target domain. This conceptually simple approach aims to mitigate the impact of domain shift and improve the model adaptability to diverse data distributions.

\begin{figure}
    \centering
    \includegraphics[width=.6\textwidth]{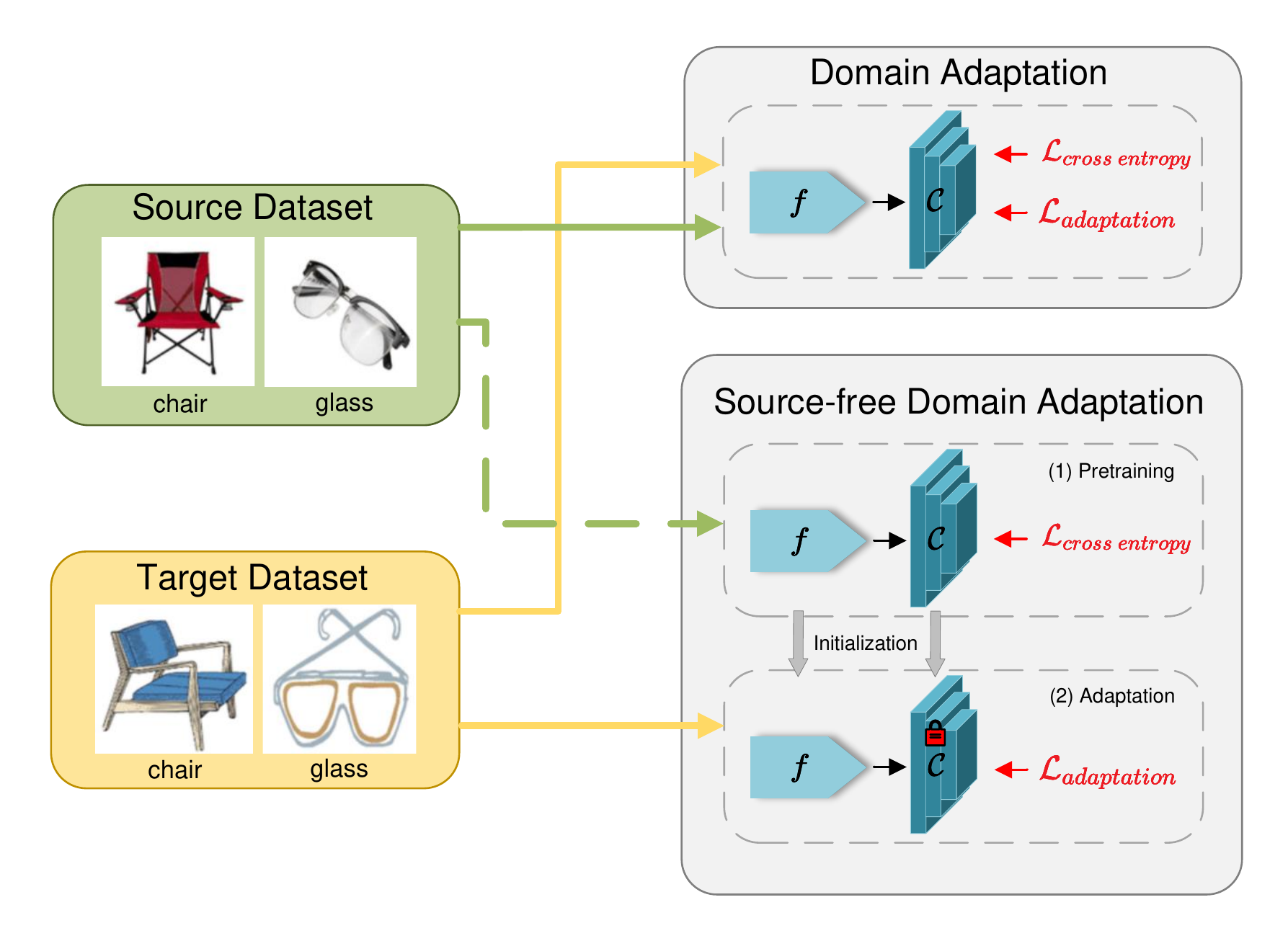}
    \caption{The different between domain adaptation and source-free domain adaptation.}
    \label{fig:sfda-da}
\end{figure}

With the growing concerns about data privacy and the bottlenecks in transferring large datasets, the coexistence of source and target data becomes highly unrealistic. In such privacy-preserving scenarios, traditional unsupervised domain adaptation (DA) methods that rely on both source and target data are not feasible. As a result, the concept of source-free domain adaptation (SFDA) has emerged to address this challenge. As shown in Fig.\ref{fig:sfda-da}, in traditional domain adaptation, both source and target domain data are used for adaptation, while in SFDA, the target domain data is the only available data for adaptation, making it a more challenging setting. Therefore, SFDA methods are designed to adapt the pretrained source model to the target domain with only the target domain data, without relying on the source domain data during the adaptation process. To achieve this, SFDA methods focus on learning domain-invariant representation or leveraging transferable knowledge from the source model to effectively adapt to the target domain without the need for source data during adaptation. Existing SFDA methods \cite{u-sfan,sfit,sfda:avatar} focus on improving the learning of domain invariant or variant representation. However, these methods often require an auxiliary network \cite{3cgan,a2net}, or involve complex additional data processing \cite{CoWA-JMDS,balancing}. Moreover, other SFDA method \cite{memory:auxiliary} is susceptible to the negative effects of noisy labels, leading to the prediction of incorrect target pseudolabels. 

The aforementioned observations serve as a motivation to address the data shift issue in SFDA. This task faces two main challenges: the absence of labeled target data and the inability to directly access source data, relying solely on the pretrained source model. In the closed set setting, where classes are shared between the source and target domains \cite{universal,universal:sfda}, it is reasonable to assume that the pretrained source model can learn class representation applicable to the target data. Consequently, even if the source and target data exhibit shifts in the feature space, the features extracted by the source model on the target data can still form rough clusters based on intrinsic class representation information (e.g., ensuring that a husky is never classified as a parrot)\cite{shot, nrc}. As a result, the softmax output of similar features should exhibit high consistency.

Inspired by the contrastive learning technique \cite{moco,infonce}, which learn feature representation by comparing positive and negative samples, we use the data itself to provide supervision for network learning. Unlike previous methods that simply added samples or modified the definition of positive pairs, we introduce two probability functions that represent the likelihood of a sample having the same category as its positive and negative samples, respectively. The negative logarithm of these two probability functions is employed as the objective function. Subsequently, we propose a novel method called Contrast and Clustering (\textbf{CaC}) that integrates these principles to perform SFDA.

Specifically, our method incorporates meticulous design of positive and negative samples, refining negative pairs pool by extended neighbors, and employs noise-contrastive estimation theory to approximate the objective function. Unlike the methodology outlined in \cite{nepos1,nepos2}, which exclusively employs neighbors for forming positive pairs, our approach takes a distinct perspective. We leverage neighbors not only for positive pairs but also for negative pairs, thereby enriching the learning process. By designating neighbors of remaining samples as negatives, our model is exposed to an extensive array of dissimilar samples. This deliberate diversification of negative pairwise relationships enhances the model's discriminative ability, enabling it to effectively differentiate between different categories. In addition, acknowledging the significance of harder negative pairs in enhancing learning efficiency\cite{neg:hard,lessCanMore}, we introduce extended neighbors to refine the negative pairs pool. This is achieved by excluding similar samples from both nearest neighbors and extended neighbors, contributing to a more robust learning process. The computation for obtaining these extended neighbors is cost-effective, as they can be queried from the existing bank directly. However, in practice, negative pairs term is not always beneficial to the results, especially when dealing with class-imbalanced datasets. To address the problem of categories imbalance, we employ the method of recursively decoupling positive and negative samples during training\cite{decoupled}. Our experimental results demonstrate the effectiveness of the proposed method on three source-free domain adaptation benchmarks.

The primary contributions of this work are as follows:
\begin{enumerate}

\item we propose an new optimization function designed to bolster the learning ability of domain-invariant features, which is achieved by constructing abundant positive and negative pairs through neighborhood data.

\item Excluding similar data from neighbors and extended neighbors refines negative pool, reducing noise and confusion in the learning process. Efficient retrieval of expanded neighbors enhances computational efficiency from the existing bank.
    
\item The experimental outcomes on three benchmark datasets validate the efficacy of our proposed method, surpassing other source-free domain adaptation approaches.
\end{enumerate}

\section{Related Work}\label{sec:related}
\subsection{Domain Adaptation}\label{subsec:related_domain}

Domain adaptation(DA) attempts to bridge the distribution gap between two domains to maximize the performance on target domain. There are four typical settings: close-set\cite{nrc}, open-set\cite{open}, partial-set\cite{partial}, and universal\cite{universal}. Among them, close-set DA is the most commonly DA setting in which the number of categories in the source and target domains is identical. Conventional close-set DA methods assume that the categories are fully shared between the source and target domains. With this assumption, a number of DA methods based on learning invariant representation have been proposed to align distributions, which can be classified into three categories: domain-level alignment methods\cite{do-level}, class-level alignment methods\cite{cl-level} and both domain- and class-level alignment methods\cite{both-level}. 

To generate similar feature distributions from different domain data, the early adversarial adaptation method \cite{dann} combines domain adaptation with a two-player game similar to generative adversarial networks(GAN). CDAN\cite{cdan} extends the conditional adversarial mechanism to enable discriminative and transferable domain adaptation. However, these GAN-based methods\cite{gragan,advgan} require auxiliary network. Without additional generative network, other methods obtain domain invariant features to minimize the domain discrepancy by conditional entropy minimization\cite{cem1,cem2,cem3} or clustering loss\cite{caco,srdc,closs1}. McDalNets\cite{cem2} formulates a new adaptation bound for DA based on the induced MCSD divergence as a measure of domain distance. CaCo\cite{caco} introduces contrastive learning, encouraging networks to learn representation with different categories but different domains. SRDC\cite{srdc} directly reveals intrinsic target discrimination by discriminative clustering of target data. However, the source data are not directly available in practice due to privacy issues, making these methods inapplicable.

\section{Related Work}\label{sec:related}
\subsection{Domain Adaptation}\label{subsec:related_domain}

Domain adaptation(DA) attempts to bridge the distribution gap between two domains to maximize the performance on target domain. There are four typical settings: close-set\cite{nrc}, open-set\cite{open}, partial-set\cite{partial}, and universal\cite{universal}. Among them, close-set DA is the most commonly DA setting in which the number of categories in the source and target domains is identical. Conventional close-set DA methods assume that the categories are fully shared between the source and target domains. With this assumption, a number of DA methods based on learning invariant representation have been proposed to align distributions, which can be classified into three categories: domain-level alignment methods\cite{do-level}, class-level alignment methods\cite{cl-level} and both domain- and class-level alignment methods\cite{both-level}. 

To generate similar feature distributions from different domain data, the early adversarial adaptation method \cite{dann} combines domain adaptation with a two-player game similar to generative adversarial networks(GAN). CDAN\cite{cdan} extends the conditional adversarial mechanism to enable discriminative and transferable domain adaptation. However, these GAN-based methods\cite{gragan,advgan} require auxiliary network. Without additional generative network, other methods obtain domain invariant features to minimize the domain discrepancy by conditional entropy minimization\cite{cem1,cem2,cem3} or clustering loss\cite{caco,srdc,closs1}. McDalNets\cite{cem2} formulates a new adaptation bound for DA based on the induced MCSD divergence as a measure of domain distance. CaCo\cite{caco} introduces contrastive learning, encouraging networks to learn representation with different categories but different domains. SRDC\cite{srdc} directly reveals intrinsic target discrimination by discriminative clustering of target data. However, the source data are not directly available in practice due to privacy issues, making these methods inapplicable.

\subsection{Source-Free Domain Adaptation} \label{subsec:related_sfda}
The above-mentioned normal domain adaptation methods need to access source domain data during target adaptation. To address this challenging unsupervised DA setting with only a trained source model provided as supervision, many methods\cite{sfda:tai,msst1,sfda:comparison,sfda:impression} have emerged to tackle source-free domain adaptation(SFDA), which has no way of accessing source data. According to application scenario, SFDA can be divided into three categories: single source to single target\cite{shot,a2net}, single source to multiple target\cite{ssmt} and multiple source to single target\cite{balancing,msst2}. Undoubtedly, the shortage of training data may increase the difficulty of model adaptation. To solve this problem, existing studies\cite{shot,3cgan,nrc,sfda:avatar,u-sfan} have pay more attention to single source to single target. SHOT\cite{shot} tunes the source classifier to encourage interclass feature clustering by maximizing mutual information and pseudolabeling. 3C-GAN\cite{3cgan} is based on conditional GAN to provide supervised adaptation by regularizing the source domain information gradually. NRC\cite{nrc} proposes neighborhood clustering, which performs predictive consistency among local neighborhoods. CPGA\cite{sfda:avatar} proposes a contrastive prototype generation strategy to generate feature prototypes for each class. U-SFAN\cite{u-sfan} accounts for uncertainty by placing priors on the parameters of the source model.

Compared to other methods, our approach differs in that it indirectly learns domain-invariant features through feature generation, such as GAN (Generative Adversarial Network) \cite{3cgan}, generation strategies \cite{sfda:avatar}, or parameter update strategies \cite{dipe}. These methods can be quite computationally costly. Our method also differs from other neighbor-based approaches. We takes a distinctive approach by employing expanded neighbors not for the purpose of identifying prototype class features, as seen in previous methods like NRC \cite{nrc}, but rather for excluding similar samples from the negative pairs pool. In this way, the utilization of expanded neighbors for negative pairs exclusion helps in creating a more informative set of negative pairs. This encourages the model to learn more generalized and domain-invariant features, enhancing its adaptability and performance across different domains.

\subsection{Contrastive Learning } \label{subsec:related_contrast}

Contrastive learning(CL) is a type of self-supervised learning that maximizes the distance between different categories and minimizes the distance between the same categories. As one of the classical methods, InfoNCE\cite{infonce} is based on the principle of noise-contrast estimation\cite{nce} and converts data distribution solving into a multi-classification problem by constructing positive and negative sample pairs. There are various ways to construct positive and negative sample pairs. InvaSpread\cite{invaspread} and SimCLR\cite{simclr} construct pairs samples from the current mini-batch, where the augmented samples are positive pairs. CMC\cite{cmc} treats data from different views of the same scene as positive pairs and data from different scenes as negative pairs. NNCLR\cite{little} uses data augmentation and its nearest neighbors in the memory bank as positive sample pairs. DCL\cite{decoupled} removes positive sample pairs from the denominator in contrast loss to achieve positive and negative term decoupling. 

In domain adaptation, CL has been used for alignment of instances to address domains discrepancy\cite{cl:can,cl:swav,cl:clda,cl:cda,cl:hcl}. CAN\cite{cl:can,cl:cdd} modifies Maximum Mean Discrepancy (MMD)\cite{mmd} loss to be used as a contrastive loss. SwAV\cite{cl:swav} proposes a contrastive pre-training method for UDA. CLDA\cite{cl:clda} employs contrastive learning to form stable and correct cluster cores in the target domain. HCL\cite{cl:hcl} investigates memory-based learning for unsupervised source-free domain adaptation that learns discriminative representation for target data. However, all the above CL-based methods treat only the augmented instance of the same input as positive pairs, and the rest of data or the random augmented data in current batch-size as negative pairs. 

Compared with the above methods, our work has the advantage of setting the nearest multiple neighbors as positive pairs, which allows clustering directly in the original feature space without the use of augmentation techniques. More importantly, different from other neighbor-based contrastive learning methods\cite{little,ncl}, we remove false negative pairs to prevent class clustering from collapsing. Even though CDA\cite{cl:cda} takes this into account, it is accomplished by computing and sorting similarities, whereas our CaC is achieved by expanded neighbors without additional computation. Additionally, CDA removes only the exact same number of negative pairs in a mini-batch., while our CaC adjusts the exclusion number according to the similarity of the samples within the current mini-batch.

\section{Method}\label{sec:method}
\subsection{Problem Definition}\label{subsec:definition}
\begin{table*}[h]
    \centering
    \caption{Symbols and corresponding definitions.}
\label{tab:symbol}
    \resizebox{1.\linewidth}{!}{
    \begin{tabular}{llll}
    \toprule
    {\makebox[0.15\linewidth][l]{Symbols}}&  {\makebox[0.25\linewidth][l]{Definitions}} &  {\makebox[0.15\linewidth][l]{Symbols}}&  {\makebox[0.25\linewidth][l]{Definitions}}\\
     \hline
    $\mathcal{D}_{s},\mathcal{D}_{t}$ & The source and target domains     &$S$ & The size of mini-batch \\
    $x_{s}, x_{t}$ & The source and target sample    & $C$  & The number of the class   \\
    $m, n$ & The number of source and target sample  & $d$  &  The size of feature dimension   \\ 
    $x_{i}$ & The i-th sample in target domain & $\mathcal{K}$ & Nearest neighbors/positive pairs set of sample $x_{i}$\\
$z(x_{i})$ &  Softmax output of sample $x_{i}$ &$\mathcal{O}$ & Dissimilar samples/negative pairs set of sample $x_{i}$\\  
    $z_{l}(x_{i})$ &the $l$-th element of $z(x_{i})$ & $f$ &Feature extractor network\\
    $K$ &  The number of nearest neighbors     & $\mathcal{C}$ & Classifier network\\
   $\mathcal{B}$ & The current mini-batch set with the excluded sample $x_{i}$   &&\\
 
    \bottomrule
    \end{tabular}
    }
\end{table*}


Given the model $\mathcal{M}_{s}$ trained on the labeled source domain $\mathcal{D}_{s}=\left \{ {x}_{s} ,{y}_{s} \right \}=\left \{ {x}_{i} ,{y}_{i} \right \}_{i=1}^{m}$ and the unlabeled target domain $\mathcal{D}_{t}=\left \{ {x}_{t} ,{y}_{t} \right \} = \left \{ {x}_{i}  \right \}_{i=1}^{n}$, we assume that the feature space $ \mathcal{X}_{s} = \mathcal{X}_{t}$ and the label space $ \mathcal{Y}_{s}=  \mathcal{Y} _{t}$. However, there exists a discrepancy between the marginal probabilities marginal probability $P _{s}\left ( {x}_{s}  \right ) \ne P_{t}\left ( {x}_{t}  \right )$ with conditional probability $ \mathcal{Q} \left ( {y_{s}} \mid  {x_{s}}  \right ) \ne \mathcal{Q} \left ( {y_{t}} \mid {x_{s}}  \right )$. In this paper, the target domain and source domain have the same $C$ categories, which is known as the closed-set problem. To address this, our method splits the model $\mathcal{M}_{s}$ into two parts: a feature extractor $f$, and a classifier $\mathcal{C}$. The output of the model is denoted as $z(x_{i})=\mathcal{C}(f(x_{i}))$. For clarity, we provide a list of all symbols used in this paper along with brief explanations in Table\ref{tab:symbol}.

The primary objective of source-free domain adaptation is to train a feature extractor $f$ and a classifier $\mathcal{C}$ that can accurately predict the labels ${y}_{t}$ for input data ${x}_{t}$ in the target domain $\mathcal{D}_{t}$ without relying on access to the source data. This paper strictly adheres to the SFDA setting, ensuring that the source data remains inaccessible throughout the target model training process. Thus, the utilization of the source data is solely restricted to training the source model and is precluded during the training of the target model.

\subsection{InfoNCE Revisited}
InfoNCE is a loss function widely used for contrastive learning. It defines the augmented sample of each sample as its positive sample, and the remaining samples in current mini-batch as negative sample. This loss function is as follows:

\begin{equation}
    \mathcal{L}^{\textls[10]{InfoNCE}}=\sum_{i=1}^{n}log\frac{e^{z(x_{i})^{T}z(\tilde{x}_{i} )/\tau} } {\sum_{b \in \mathcal{B} }e^{z(x_{i})^{T}z(x_{b})/\tau}+e^{z(x_{i})^{T}z(\tilde{x}_{i})/\tau}}
\end{equation}
where $\tilde{x}$ denotes the augmented sample, $x_{b}$ denotes the remaining samples, and $\tau\in \mathcal{R^{+}}$ is a scalar temperature parameter.

\begin{figure}[h]
    \centering
    \includegraphics[width=1.\textwidth]{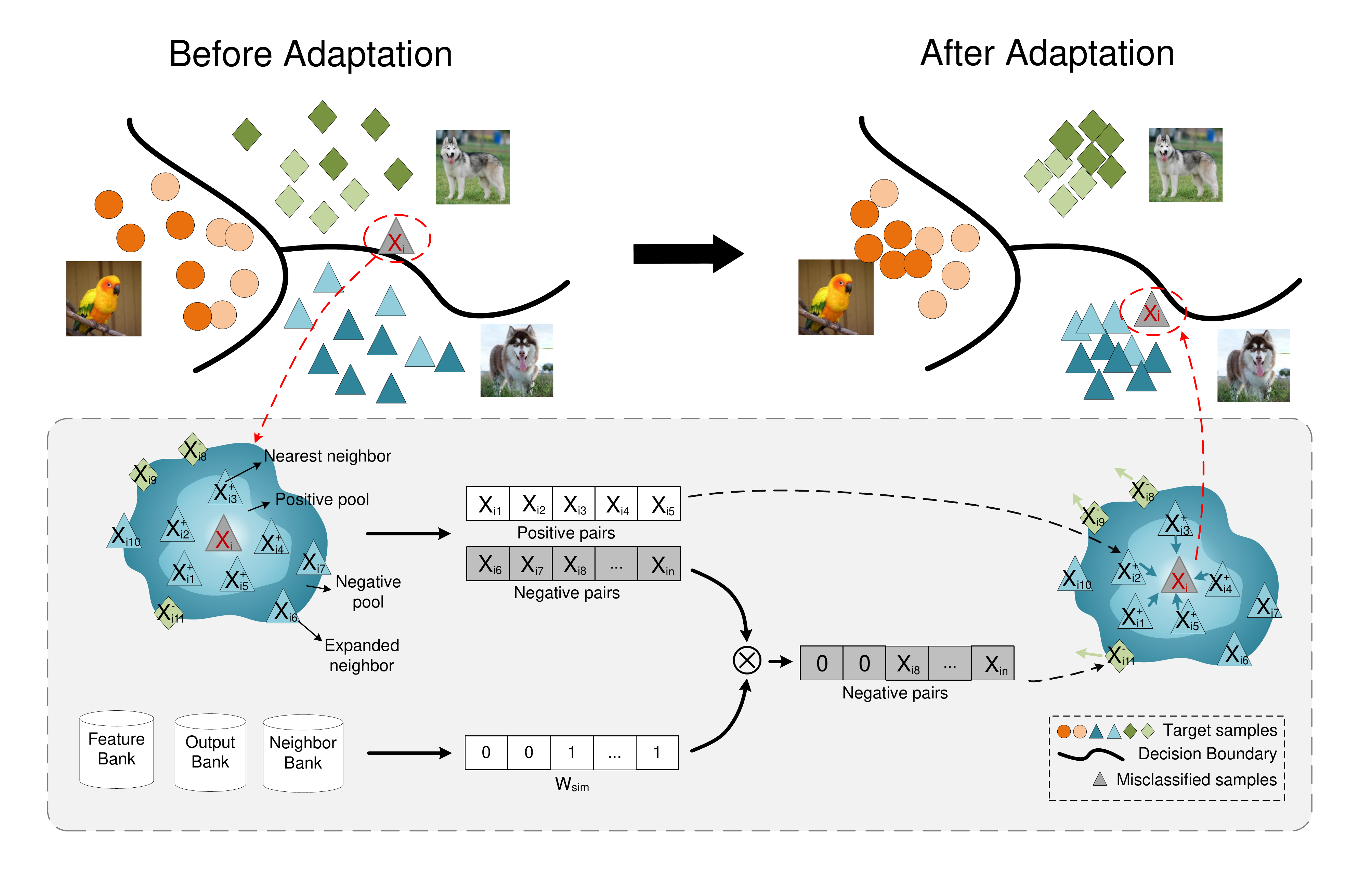}
    \caption{An overview of the proposed method. When $X_{i}$ has been misclassified, CaC is able to learns features of the unlabeled target data from pretrained source network, and cluster samples that are highly similar in the prototype feature space. $W_{sim}$ is the weight matrix, which is used to eliminate similar samples from the pool of negative pair samples.  }
    \label{fig:cac}
\end{figure}

\subsection{Motivation}\label{sec_motivation}

In contrast to previous methods like InfoNCE that consider an augmented sample as a positive pairs, we define the positive pairs for each instance $x_{i}$ as its neighbor samples, specifically the top-K similar samples in the feature embedding. This enables us to directly perform clustering on the data without relying on any generative techniques and allows for a larger number of positive pairs. The positive pairs are selected from the k-nearest neighbor set $\mathcal{K}$ of instance $x_{i}$, based on either cosine similarity or Euclidean distance. The remaining samples not in this set are chosen as the negative pairs. This setup allows for the expansion of InfoNCE to include multiple positive pairs, and the loss function $\mathcal{L}$ can be defined as follows:

\begin{equation}
    \mathcal{L} =-\sum_{i=1}^{n}\sum_{j \in \mathcal{K}}log\frac{ e^{z(x_{i})^{T}z(x_{j}) } } {\sum_{b \in \mathcal{B}, b\ne j}e^{ z(x_{i})^{T}z(x_{b})}+ e^{z(x_{i})^{T}z(x_{j})}}
\end{equation}

Intuitively, in scenarios where similar samples (e.g., husky and Alaskan malamute) belong to different categories, the effectiveness of the above loss function is limited. It merely extends InfoNCE to incorporate multiple positive pairs, thereby promoting the network to classify samples with high similarity into the same category and those with low similarity into different categories. However, when two categories exhibit similar features, the risk of misclassification arises. The model may erroneously pull closer to neighbors from different categories, leading to suboptimal performance. To overcome this limitation and ensure more effective discrimination, the network needs to encourage closer proximity among neighbors belonging to the same category while simultaneously pushing away neighbors from different categories. By doing so, the model can better distinguish between similar but distinct categories, enhancing the overall discriminative power of the learned features and mitigating the potential misclassification issue.

\subsection{Proposed Method}

Assuming that the target features learned by the source pretrained feature extractor exhibits the ability to form clusters\cite{shot,nrc}, our proposed method, CaC, harnesses this intrinsic characteristic of the pretrained model to conduct SFDA by effectively incorporating neighborhood information. Our method is illustrated in Fig.\ref{fig:cac}. 

In a C-class classification scenario, according to the softmax function, the probability of sample ${x}_{i}$ belonging to the $l$-th class is expressed as follows:
\begin{equation}\label{prob}
P(Y=l \vert X=x_{i})= \frac{e^{z_{l}(x_{i})}}{\sum_{c\in C} {e^{z_{c}(x_{i})}} }
\end{equation}
where $z(x_{i})$ is the logits that produced by source model. $z_{l}(x_{i})$ represents the $l$-th element in $z(x_{i})$ and it can be interpreted as the probability that instance ${x}_{i}$ belongs to the $l$-th class. This probability calculation is performed using the softmax function, which ensures that the probabilities for all classes sum up to 1 for each sample.

We now proceed to examine the following conditions. Given an unlabeled sample $x_{i}$ and it belongs to $l$-th class, its k-nearest neighbor set is denoted as $\mathcal{K} =\{  x_{i^{1}},x_{i^{2}},...,x_{i^{k}}|M(f(x_{i^{h}}),f(x_{i}))>M(f(x_{q}),f(x_{i})),q\ne i , q \ne i^{h},h=1,2,...,k\}$, where $M$ represents a distance metric or a similarity measure between two feature embeddings. Additionally, the set $\mathcal{O}$ represents the collection of other samples in the mini-batch. The methods used for finding $\mathcal{K}$ and $\mathcal{O}$ are thoroughly described in Sec.\ref{subsec:find} and Sec.\ref{subsec:expanded}, respectively. Intuitively, it is expected that $x_{i}$ and its neighbor set $\mathcal{K}$ belong to the same class, indicating that their output probabilities of class membership are highly consistent. With the Eq.\eqref{prob}, the probability of each sample $x_{j}\in \mathcal{K}$ belonging to  $l$-th class should higher than samples in $\mathcal{O}$. Therefore, With the definition of $\mathcal{K}$, we claim that the Proposition \ref{pro:class consist} holds:

\begin{proposition}[Class consistency of positive pairs]\label{pro:class consist}
$\forall x_{j}\in \mathcal{K}$, $\forall x_{q}\in \mathcal{D}_{t},  q\ne j , q\ne i$, $ e^{z(x_{i})^{T} z_{l}(x_{j})} >e^{z(x_{i})^{T} z_{l}(x_{q})}$

\end{proposition}

According to Proposition \ref{pro:class consist}, when considering the sample $x_{i}$ and its k-nearest neighbor set $\mathcal{K}$, their output should exhibit a higher level of similarity compared to the output of the k-nearest neighbor set $\mathcal{O}^{k}$ containing the remaining data in the current batch. This similarity is referred to as the $"$output consistency of the positive pairs."

\begin{proposition}[Output consistency of positive pairs]\label{pro:consist}
$\forall x_{j}\in \mathcal{K}$, $\forall x_{q}\in \mathcal{D}_{t} ,  q\ne j , q\ne i$, $ e^{z(x_{i})^{T} z(x_{j})} >e^{z(x_{i})^{T} z(x_{q})}$
 
\end{proposition}

We define two likelihood functions, $P_{i,j}^{same}$ and $P_{i,j}^{dis}$, as follows: $P_{i,j}^{same}$ represents the probability that ${x}_{i}$ and its positive samples belong to the same category. Conversely, $P_{i,j}^{dis}$ represents the probability that ${x}_{i}$ and the neighbors of its corresponding negative samples share the same category. These functions quantify the relationships between ${x}_{i}$ and its positive and negative samples in terms of class membership probabilities.

\begin{equation}
\label{eq_same}
  P_{i,j}^{same}= \prod_{x_{j}\in \mathcal{K}}\frac{e^{z(x_{i})^{T} z(x_{j})} }{\sum_{q\ne i}e^{z(x_{i})^{T}z(x_{q})} } 
\quad \quad \quad
    P_{i,j}^{dis}= \prod_{x_{j}\in \mathcal{O}^{k}}\frac{e^{z(x_{i})^{T} z(x_{j})} }{\sum_{q\ne i}e^{z(x_{i})^{T}z(x_{q})} } 
\end{equation}
where $\mathcal{O}^{k}$ denotes the corresponding k-nearest neighbors of $\mathcal{O}$. As the output consistency of neighbors is desired to be maximized, we propose to minimize the following negative logarithmic objective function and finally achieve clustering.
\begin{equation}
\begin{split}
\mathcal{L} &= -\frac{1}{n}  \sum_{i=1}^{n} log\frac{P_{i,j}^{same}}{P_{i,j}^{dis}}
\\&= \frac{1}{n} \sum_{i=1}^{n} (\sum_{x_{j}\in \mathcal{O}^{k}}{z(x_{i})^{T} z(x_{j})} - \sum_{x_{j}\in \mathcal{K}} z(x_{i})^{T} z(x_{j}))+ \frac{1}{n} \sum_{i=1}^{n}\left | \mathcal{K}  \right | (1-\left |\mathcal{O} \right | )log\sum_{q \ne i} e^{z(x_{i})^{T}\cdot z(x_{q})} 
\end{split}
\end{equation}

Step 1 (upper bound): we establish an upper bound for $\mathcal{L}$ using the set sizes $\left |\mathcal{O} \right |=\left | \mathcal{K} \right | \times \left | \mathcal{B} \right | $. Additionally, the constant term $\left | \mathcal{K} \right | (1-\left |\mathcal{O} \right |) < 0$ and make use of properties related to the convexity of the logarithm function, along with Jensen's inequality:$f(\frac{\sum_{i=1}^{n} x_{i}}{n} )\ge  \frac{\sum_{i=1}^{n} f(x_{i})}{n}  $. As a result, we arrive at the following expression:

\begin{equation}
\label{eq:upper}
\begin{split}
\mathcal{L} \le & \frac{1}{n} \sum_{i=1}^{n} (\sum_{x_{j}\in \mathcal{O}^{k}}{z(x_{i})^{T} z(x_{j})} - \sum_{x_{j}\in \mathcal{K}} z(x_{i})^{T} z(x_{j})) \\
&+\frac{1}{n} \sum_{i=1}^{n} \left | \mathcal{K}  \right | (1-\left |\mathcal{O} \right |)( \frac{1 }{n-1} \sum_{q \ne i} z(x_{i})^{T}\cdot z(x_{q})+log(n-1))
\end{split}
\end{equation}

Step 2 (approximation): in real-world scenarios, calculating the last term in Eq.\eqref{eq:upper} becomes computationally challenging due to the large size of the dataset. However, we can overcome this computational burden by leveraging the success of noise-contrastive estimation theory\cite{infonce,nce}. By approximating the entire dataset using negative pairs, we can obtain the following simplified expression for the function.

\begin{equation}\label{eq:app}
\begin{split}
\mathcal{L} \approx & \frac{1}{n} \sum_{i=1}^{n} (\sum_{x_{j}\in \mathcal{O}^{k}}{z(x_{i})^{T} z(x_{j})} - \sum_{x_{j}\in \mathcal{K}} z(x_{i})^{T} z(x_{j})) \\
&+\frac{1}{n} \sum_{i=1}^{n} \left | \mathcal{K}  \right | (1-\left |\mathcal{O} \right |)( \frac{1 }{n-1} \sum_{x_{j}\in \mathcal{O}^{k}} z(x_{i})^{T}\cdot z(x_{j})+log(n-1))
\end{split}
\end{equation}

Finally, we omit the constant term in Eq.\eqref{eq:app} and obtain
\begin{equation}\label{eq_cac}
\mathcal{L}^{\mathrm{CaC}}=\frac{1}{n} \sum_{i=1}^{n} (\underbrace{ \sum_{x_{j}\in \mathcal{O}^{k}}{z(x_{i})^{T} z(x_{j})}}_{neg: negative\ pairs} - \underbrace{ \sum_{x_{j}\in \mathcal{K}} z(x_{i})^{T} z(x_{j})}_{pos: positive\ pairs})
\end{equation}

The final function denoted as CaC, which is the approximation of the original objective function. To illustrate the impact of these two terms, we conduct the experiment in the ablation experiment.

\subsection{Finding the Nearest Neighbors}
\label{subsec:find}

To retrieve the nearest neighbors for batch training, we create three memory banks: $\mathcal{F} \in \mathbb{R}^{n \times d}$ stores all target features, $\mathcal{P} \in \mathbb{R}^{n \times C}$ stores the corresponding prediction scores, and $\mathcal{N} \in \mathbb{R}^{n \times K}$ stores the corresponding top-K nearest data. These memory banks are initialized with all target features and their predictions, and only the features and predictions computed in each mini-batch are used to update them, following a similar approach as in prior studies \cite{memory:auxiliary,nrc}.

For each sample $x_{i}$, its nearest neighbor set, denoted as $\mathcal{K}=topK(f(x_{i}))$, consists of samples with the top-K highest similarity to the memory bank $\mathcal{F}$. This set is utilized to compute the positive pairs in Eq.\eqref{eq_cac}. The cosine similarity is chosen as the distance measurement in this work. The similarity between two samples is maximized when their softmax outputs have the same prediction class and are close to a one-hot vector.

\subsection{Finding the Truly Negative Pairs}
\label{subsec:expanded}
In Eq.\eqref{eq_cac}, the negative pairs for a sample $x_{i}$ are the remaining samples in the mini-batch, excluding the positive pairs. A critical consideration is that these remaining samples in the mini-batch might belong to the same category as $x_{i}$. These similar samples should be omitted from the corresponding negative pairs set $\mathcal{O}$ because their inclusion could introduce noise and confusion into the learning process. The genuinely dissimilar pairs for a sample are identified by excluding similar samples found in both neighbors and extended neighbors. 

The concept of extended neighbors in this method differs from the objective in NRC\cite{nrc}, where the goal is to gather more information about similar samples to achieve clustering. In our approach, the emphasis is on refining negative pairs by incorporating expanded neighbor relationships to enhance sample discrimination and feature learning. For instance, the sample $x_{j}$ is considered a similar sample of $x_{i}$ if $x_{j}$ is among the top-K nearest neighbors of the top-K nearest neighbors of $x_{i}$, i.e., $x_{j} \in topK(topK(f(x_{i})))$. This approach is efficient and does not require any additional computation, as the nearest neighbors are already computed and update the corresponding indexes for each sample in the neighbor bank $\mathcal{N}$. Therefore, we can directly retrieve their expanded neighbors from the bank $\mathcal{N}$.

To achieve this, we employ a weight matrix $W_{sim} \in \mathbb{R}^{S \times S}$ to exclude similar samples, where a zero element indicates that the sample at that location has been masked. If $x_{j}$ is a positive sample to $x_{i}$, the j-th column of the i-th row in $W_{sim}$ is set to zero, and the other positions are set to one. Therefore, the negative pairs set for each sample $x_{i}$ is denoted as:
\begin{equation}
\label{eq_weight}
\mathcal{O} = nondiag(W_{sim})[:,i] \cdot \mathcal{B}
\end{equation}
where $nondiag(W_{sim}) \in \mathbb{R}^{S-1 \times S}$ is the operation to obtain the non-diagonal elements of $W_{sim}$, and the set $\mathcal{B} \in \mathbb{R}^{(S-1) \times 1}$ represents the remaining samples in the current mini-batch.

\begin{algorithm}[tb]
    \caption{Learning Nearest Pair Representation for SFDA}
    \label{alg:algorithm}
    \textbf{Input}: Source-pretrained model $\mathcal{M}_{s}$ and unlabeled target sample $\mathcal{D}_{t}$.
    \begin{algorithmic}[1] 
        \STATE 
        Build three memory banks to store all the target features( $\mathcal{F}$) and predictions($\mathcal{P}$) and the indexes of neighbors($\mathcal{N}$).
        \STATE Initialize the $\mathcal{P}$ and $\mathcal{F}$ banks by model $\mathcal{M}_{s}$.
        \WHILE{training}
        \STATE Sample a mini-batch $\mathcal{T}$ from $\mathcal{D}_{t}$ and update memory banks $\mathcal{P}$ and $\mathcal{F}$.
        \STATE For each feature in $\mathcal{T}$, find its K-nearest neighbors $\mathcal{K}$ and update memory bank $\mathcal{N}$.
        \STATE Retrieve expanded neighbors from memory bank $\mathcal{N}$ to generate $W_{sim}$
        \STATE Compute the loss function $\mathcal{L}^{\mathrm{CaC}}$
        \STATE Back-propagate with the loss function and update the network parameters
        \ENDWHILE
        \RETURN solution
    \end{algorithmic}
\end{algorithm}

In conclusion, the two terms in our algorithm interact to achieve self-supervision of the features. Positive pairs enhance the consistency of outputs, while negative pairs enhance the diversity of outputs. The weight matrix $W_{sim}$ plays a crucial role in mining truly hard negative pairs. The overall algorithm is presented in Algorithm~\ref{alg:algorithm}.

\section{Experiments}\label{sec:exp}
\subsection{Datasets}

We conduct the experiments on three benchmark
datasets: 
{\bf VisDA} is a more challenging dataset, with 12-class synthetic-to-real object recognition tasks. Its source domain consists of 152k synthetic images while the target domain contains 55k real object images. 
{\bf Office-Home} contains 4 domains(Art, Clipart, Real World, Product) with 65 classes and a total of 15,500 images.
{\bf Office-31} contains 3 domains(Amazon, Webcam, DSLR) with 31 classes and 4652 images. 

\subsection{Evaluation}
The column SF in the tables denotes source-free setting. For VisDA, we show accuracy for all classes and average over those classes (Avg in the tables). For Office-31 and Office-Home, we show the results of each task and the average accuracy over all tasks (Avg in the tables). 
\subsection{Baselines}
We compare CaC with three types of baselines: (1) source-only: ResNet\cite{resnet}; (2) UDA with source data: DANN\cite{dann}, CDAN\cite{cdan}, SRDC\cite{srdc}, CaCo\cite{caco}; and (3) source-free UDA: SHOT\cite{shot}, 3C-GAN\cite{3cgan}, NRC\cite{nrc}, CPGA\cite{sfda:avatar}, U-SFAN+\cite{u-sfan}.

\subsection{Implementation details.}\label{subsec:imple}
To ensure fair comparison with related methods, we use the same network architecture as SHOT and adopt SGD with momentum 0.9 and batch size of 64 for all datasets. Specifically, we adopt the backbone of ResNet50 for Office-Home and Office31, and ResNet101 for VisDA. The learning rate for Office-Home and  Office31 is set to 1e-3 for all layers, except for the last two newly added fc layers, where we apply 1e-2. Learning rates are set 10 times smaller for VisDA. We train 15 epochs for VisDA, 40 epochs for Office-Home and 100 epochs for Office-31. The code is available at https://github.com/yukilulu/CaC.
\begin{table*}
    \centering
     \caption{Classification accuracy (\%) on VisDA(Synthesis $\to$ Real) based on ResNet101.}
    \label{tab:VisDA}
    \resizebox{1.0\columnwidth}{!}{
    \begin{tabular}{c|c|ccccccccccccc}
    \toprule 
    \multirow{2}{*}{Method} &  \multirow{2}{*}{\makebox[0.015\textwidth][c]{SF}}  &
    \multicolumn{13}{c}{VisDA}\\
   \cline{3-15}
  & & \makebox[0.02\textwidth][c]{plane} & \makebox[0.02\textwidth][c]{bicycle} & \makebox[0.02\textwidth][c]{bus}& \makebox[0.02\textwidth][c]{car}& \makebox[0.02\textwidth][c]{horse} & \makebox[0.02\textwidth][c]{knife}& \makebox[0.02\textwidth][c]{mcycl} & \makebox[0.02\textwidth][c]{person} & \makebox[0.02\textwidth][c]{plant} & \makebox[0.02\textwidth][c]{sktbrd} & \makebox[0.02\textwidth][c]{train} & \makebox[0.02\textwidth][c]{truck} &\textbf{ Avg}\\
     \hline 
ResNet-101\cite{resnet} &\scalebox{0.75}{\usym{2613}}&55.1&53.3&61.9&59.1&80.6&17.9&79.7&31.2&81.0 &26.5 &73.5 &8.5 &52.4\\
DANN\cite{dann} &\scalebox{0.75}{\usym{2613}}& 81.9&77.7&82.8 &44.3 &81.2 &29.5 &65.1 &28.6 &51.9 &54.6 &82.8 &7.8 &57.4\\
CDAN\cite{cdan}&\scalebox{0.75}{\usym{2613}}&85.2 &66.9 &83.0 &50.8 &84.2 &74.9 &88.1 &74.5 &83.4 &76.0 &81.9 &38.0 &73.9\\
CaCo\cite{caco} & \scalebox{0.75}{\usym{2613}} & 90.4 &80.7 &78.8 &57.0 &88.9 &87.0 &81.3 &79.4 &88.7 &88.1 &86.8 &63.9 &80.9 \\
   
     \hline  \hline
SHOT\cite{shot} & $\checkmark$&94.3 &88.5 &80.1 &57.3 &93.1 &94.9 &80.7 &80.3 &91.5 &89.1 &86.3 &58.2 &82.9\\
3C-GAN\cite{3cgan} &$\checkmark$&94.8 &73.4 &68.8 &74.8 &93.1 &95.4 &88.6 &\textbf{84.7} &89.1 &84.7 &83.5 &48.1 &81.6\\

NRC\cite{nrc} & $\checkmark$ &96.8 &\textbf{91.3} &82.4 &62.4 &96.2 &95.9& 86.1 &80.6 &94.8 &\textbf{94.1} &90.4 &59.7 &\underline{85.9}  \\
CPGA\cite{sfda:avatar} &$\checkmark$ & 94.8&83.6 &79.7 &65.1 &92.5 &94.7 &90.1 &82.4& 88.8 &88.0 &88.9 &60.1 &84.1\\
U-SFAN+\cite{u-sfan} & $\checkmark$ & 94.9 &87.4 &78.0 &56.4 &93.8 &95.1 &80.5 &79.9 &90.1 &90.1 &85.3 &60.4 &82.7\\
CaC(Ours) & $\checkmark$ &\textbf{96.9} &91.0 &\textbf{83.3} &\textbf{72.3} &\textbf{96.9} &\textbf{96.1} &\textbf{90.7} &81.6 &\textbf{95.1} &92.9 &\textbf{92.0} &\textbf{63.2} & \textbf{87.7} \\
       \bottomrule
       \end{tabular}
   }
\end{table*}

\begin{table*}
\centering
\caption{Classification accuracy (\%) on Office-Home based on ResNet50.}
    \label{tab:home}
\resizebox{1.0\columnwidth}{!}{
    \begin{tabular}{c|c|ccccccccccccc}
    \toprule
    \multirow{2}{*}{Method} &  \multirow{2}{*}{\makebox[0.015\textwidth][c]{SF}}  &
    \multicolumn{12}{c}{Office-Home}\\
    \cline{3-15}
  & & \makebox[0.02\textwidth][c]{A$\to$C} & \makebox[0.02\textwidth][c]{A$\to$P} & \makebox[0.02\textwidth][c]{A$\to$R} & \makebox[0.02\textwidth][c]{C$\to$A} & \makebox[0.02\textwidth][c]{C$\to$P} & \makebox[0.02\textwidth][c]{C$\to$R} & \makebox[0.02\textwidth][c]{P$\to$A} & \makebox[0.02\textwidth][c]{P$\to$C} & \makebox[0.02\textwidth][c]{P$\to$R} & \makebox[0.02\textwidth][c]{R$\to$A} & \makebox[0.02\textwidth][c]{R$\to$C} & \makebox[0.02\textwidth][c]{R$\to$P} &  \makebox[0.02\textwidth][c]{\textbf{Avg}}\\
     \hline
ResNet-50\cite{resnet} &\scalebox{0.75}{\usym{2613}}&34.9 &50.0 &58.0 &37.4 &41.9 &46.2 &38.5 &31.2 &60.4 &53.9 &41.2 &59.9 &46.1\\
DANN\cite{dann} &\scalebox{0.75}{\usym{2613}}& 45.6 &59.3 &70.1 &47.0 &58.5 &60.9 &46.1 &43.7 &68.5 &63.2 &51.8 &76.8 &57.6\\
CDAN\cite{cdan} &\scalebox{0.75}{\usym{2613}}&50.7 &70.6 &76.0 &57.6 &70.0 &70.0 &57.4 &50.9 &77.3 &70.9 &56.7 &81.6 &65.8\\
SRDC\cite{srdc} &\scalebox{0.75}{\usym{2613}}&52.3 &76.3 &81.0 &69.5 &76.2 &78.0 &68.7 &53.8 &81.7 &76.3 &57.1 &85.0 &71.3\\
    
     \hline \hline
SHOT\cite{shot} & $\checkmark$ &57.1&78.1&81.5&68.0&78.2&78.1&\textbf{67.4}&54.9&82.2&73.3&58.8 &84.3 &71.8\\
NRC\cite{nrc}&$\checkmark$& 57.7&\textbf{80.3} &\textbf{82.0} &\textbf{68.1} &\textbf{79.8} &78.6 &65.3 &56.4 &\textbf{83.0} &71.0 &58.6 &85.6 &\underline{72.2}\\
U-SFAN+ \cite{u-sfan}&$\checkmark$&57.8 &77.8 &81.6 &67.9 &77.3 &79.2 &67.2 &54.7 &81.2 &73.3 &\textbf{60.3} &83.9 &71.9\\
CaC(Ours)  &$\checkmark$ &\textbf{59.0} &79.5 & \textbf{82.0} &67.6 &79.2 &\textbf{79.5} &66.7 &\textbf{56.5} &81.3 &\textbf{74.2} &58.3 &\textbf{84.7} & \textbf{72.4}     \\

    \bottomrule
    \end{tabular}
    }
\end{table*}

\begin{table}
\centering
\caption{Classification accuracy (\%) on Office-31 based on ResNet50.}
\label{tab:office31}
\begin{tabular}{c|c|cccccccc}
\toprule
\multirow{2}{*}{Method} &  \multirow{2}{*}{\makebox[0.015\textwidth][c]{SF}}  &
\multicolumn{7}{c}{Office-31}\\
\cline{3-9}
  & & \makebox[0.052\textwidth][c]{A$\to$D} & \makebox[0.02\textwidth][c]{A$\to$W} & \makebox[0.02\textwidth][c]{D$\to$A} & \makebox[0.02\textwidth][c]{D$\to$W} & \makebox[0.02\textwidth][c]{W$\to$A} & \makebox[0.02\textwidth][c]{W$\to$D} & \makebox[0.02\textwidth][c]{\textbf{Avg}}\\
     \hline 
ResNet-50\cite{resnet} &\scalebox{0.75}{\usym{2613}}&68.9 &68.4 &62.5 &96.7 &60.7 &99.3 &76.1\\
DANN\cite{dann}&\scalebox{0.75}{\usym{2613}}& 79.7 &82.0 &68.2 &96.9 &67.4 &99.1 &82.2  \\
CDAN\cite{cdan}&\scalebox{0.75}{\usym{2613}}& 92.9 &94.1 &71.0 &98.6 &69.3 &100.0 &87.7\\
CaCo\cite{caco} &\scalebox{0.75}{\usym{2613}}&89.7 &98.4 &100.0 &91.7 &73.1 &72.8 &87.6\\
     \hline\hline
SHOT\cite{shot} &$\checkmark$  & 94.0&90.1&74.7&98.4&74.3&99.9&88.6   \\
3C-GAN\cite{3cgan} &$\checkmark$& 92.7&93.7&\textbf{75.3} &98.5 &\textbf{77.8} &99.8&\underline{89.6}\\
NRC \cite{nrc}&$\checkmark$  &\textbf{96.0} &90.8 &\textbf{75.3} &99.0& 75.0& \textbf{100.0} &89.4\\ 
U-SFAN+ \cite{u-sfan}&$\checkmark$ &94.2 &92.8 &74.6 &98.0 &74.4 &99.0 &88.8\\
CaC(Ours) &$\checkmark$ & 95.2 & \textbf{93.8} & 74.7 & \textbf{99.1} &76.3 & 99.8 & \textbf{89.9} \\
       \bottomrule
       \end{tabular}
\end{table}

\subsection{Qualitative results}\label{subsec:main_res}
In this section, we compare our proposed CaC with the state-of-the-art methods on three DA benchmarks. In Table \ref{tab:VisDA}, Table\ref{tab:home} and Table \ref{tab:office31}, the top part shows the results for the DA methods with access to source data during adaptation. The bottom shows the results for the SFDA methods. We mark the best and second-best results in SFDA task, the best results are bolded, and the second-best results are underlined.

Specifically, CaC outperforms other state-of-the-art (SOTA) methods on the more challenging dataset \textbf{VisDA}, achieving the best results in various categories and ultimately obtaining excellent outcomes, as shown in Table \ref{tab:VisDA}. For \textbf{Office-Home}, the proposed CaC obtains better results compared to other SFDA methods, as shown in Table \ref{tab:home}. Note that our method is superior in the tasks A→C, A→R, C→R, and P→C. In addition, CaC achieves similar results to the SOTA in \textbf{Office-31}, as shown in Table \ref{tab:office31}. The main takeaway is that VisDA provides sufficient data for learning positive and negative pairs, enabling CaC to acquire better domain-invariant representation for achieving clustering of same-class samples. Moreover, CaC outperforms recent methods with source data (e.g., CaCo and SRDC), demonstrating the superiority of our proposed approach.



\subsection{ Ablation studies} \label{subsec:ablation}

\subsubsection{Ablation study on the proposed $\mathcal{L}^{\mathrm{CaC}}$}
\begin{table}
    \centering
    \caption{Classification accuracy(\%)  comparison with different components on VisDA.}
    \label{tab:each}
    \begin{tabular}{ccc|c}
    \toprule
     \makebox[0.075\textwidth][c]{$\mathcal{L}_{neg}$} &  \makebox[0.075\textwidth][c]{$\mathcal{L}_{pos}$} & \makebox[0.075\textwidth][c]{$\mathcal{W}_{sim}$} & \makebox[0.075\textwidth][c]{Avg}\\
    \hline
  $\checkmark$ &   & &  62.53\\
    & $\checkmark$& &  75.66\\
   $\checkmark$  & $\checkmark$ & &79.77 \\
   $\checkmark$  & $\checkmark$ & $\checkmark$ &\textbf{81.01} \\
       \bottomrule
       \end{tabular}
\end{table}
To investigate the improvement bring by the each component, we show the quantitative results in Table \ref{tab:each}. Note that we focus on the effects of positive and negative pairs in this experiment. Our proposed method contains two terms and one weight matrix. From this table, we can see the positive term can bring more improvement than the negative term, because the clustering ability of the positive term. While depending solely on positive term might yield lower scores, it is crucial to understand that excluding negative term can pose challenges for the model in accurately classifying complex samples. The absence of negative term can impair the model ability to differentiate certain intricate instances. These intricate samples, which might share similarities with other classes or exhibit nuanced characteristics, could undergo misclassification or cause confusion if negative term is not taken into account. Consequently, the integration of both positive and negative term becomes a critical component of effective adaptation learning. Moreover, we obtained the best performance when extracting more valuable negative sample pairs by using the weights $W_{sim}$ generated from the nearest neighbors and extended neighbors. The improvement is attributed to the utilization of weight matrix, which further mines hard samples. 

By incorporating these components, the model attains a comprehensive grasp of the subtleties and variations inherent in the dataset, thereby enhancing its capability to adeptly classify a diverse range of cases, ranging from the straightforward to the intricate.

\subsubsection{The impact of negative pairs}

\begin{figure}
\centering
\label{fig:alpha}\includegraphics[width=0.5\textwidth]{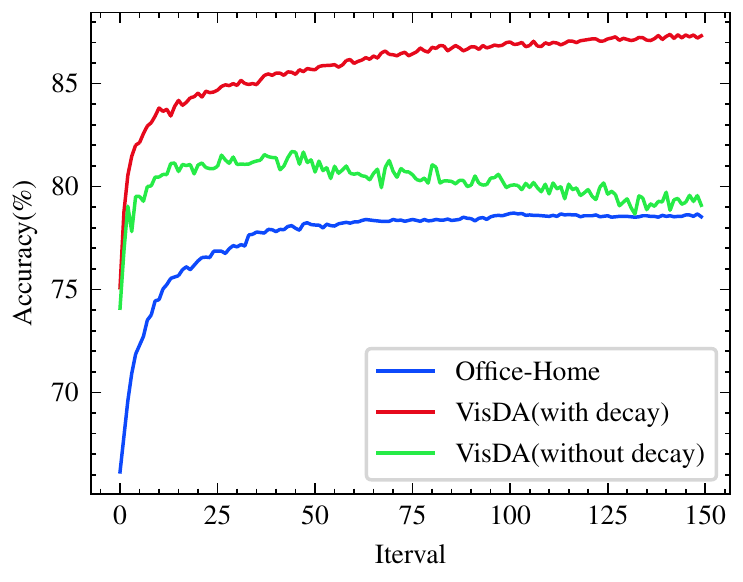}
\caption{Classification accuracy(\%)  comparisons on the Office-Home and VisDA datasets. VisDA is set to with and without decay and Office-Home is only set to without decay for a clearer comparison.}
\label{fig:acc}
\end{figure}

\begin{figure}[htbp]
\centering
\includegraphics[width=\textwidth]{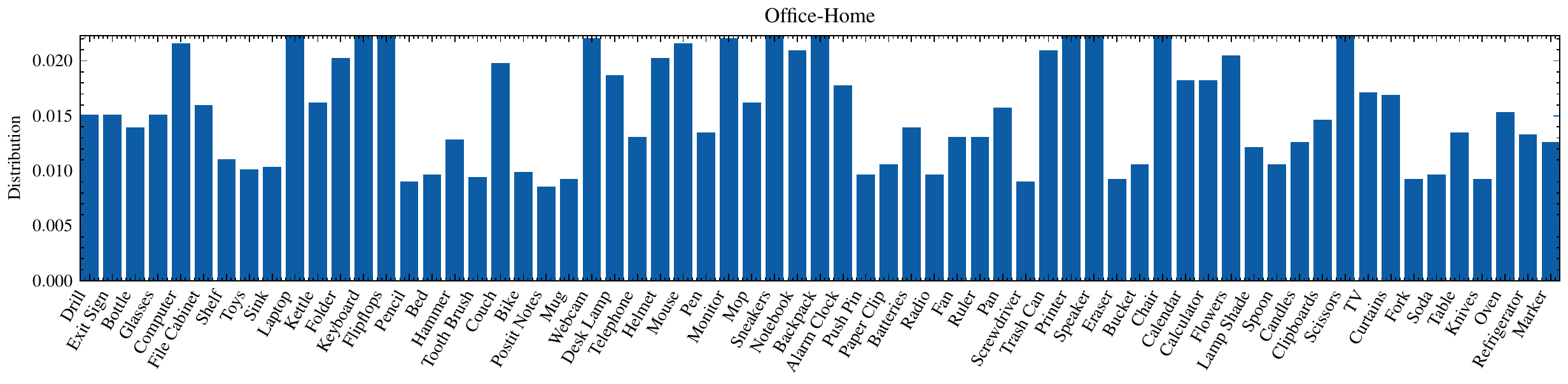}
\caption{The class distribution on Office-Home.}
\label{fig:oh class}
\end{figure}

\begin{figure}[htbp]
\centering
\includegraphics[width=0.5\textwidth]{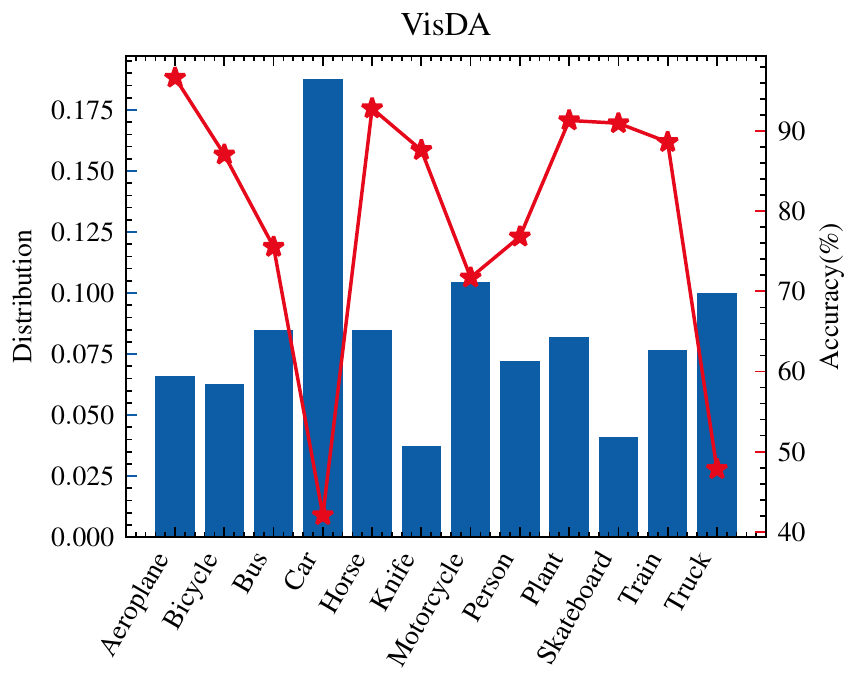}
\caption{The distribution and the classification accuracy(\%) of each class on VisDA.}
\label{fig:visda class}
\label{fig:acc each visda}
\end{figure}

We notice that the above ablation study has relative low performance in negative pairs. Therefore, we observe the result on two datasets with different sizes to further analyse the impact of negative pairs. From the result, CaC maintains accuracy improvement on Office-Home but experiences degradation at a later stage on VisDA, as shown by the green curve in Fig.\ref{fig:acc}. We first consider the class comparison between these two datasets. As illustrated in Fig.\ref{fig:oh class} and Fig.\ref{fig:visda class}, VisDA suffers from a more severe class imbalance problem than Office-Home. VisDA has a smaller number of classes, and even worse, there is a significant gap in class proportions. For instance, taking the fourth class of VisDA, "car" as an example, a large proportion of the samples in a mini-batch belong to the "car" class, causing the contrast loss to treat the other samples in the mini-batch as potentially negative pairs. As a result, the network ends up separating samples that belong to the same class, leading to a decrease in accuracy. Fig.\ref{fig:acc each visda} demonstrates that CaC is notably less accurate for classes with large quantities, such as "car" and "truck," compared to other classes.


\begin{figure}[htbp]
\centering
\includegraphics[width = .6\linewidth]{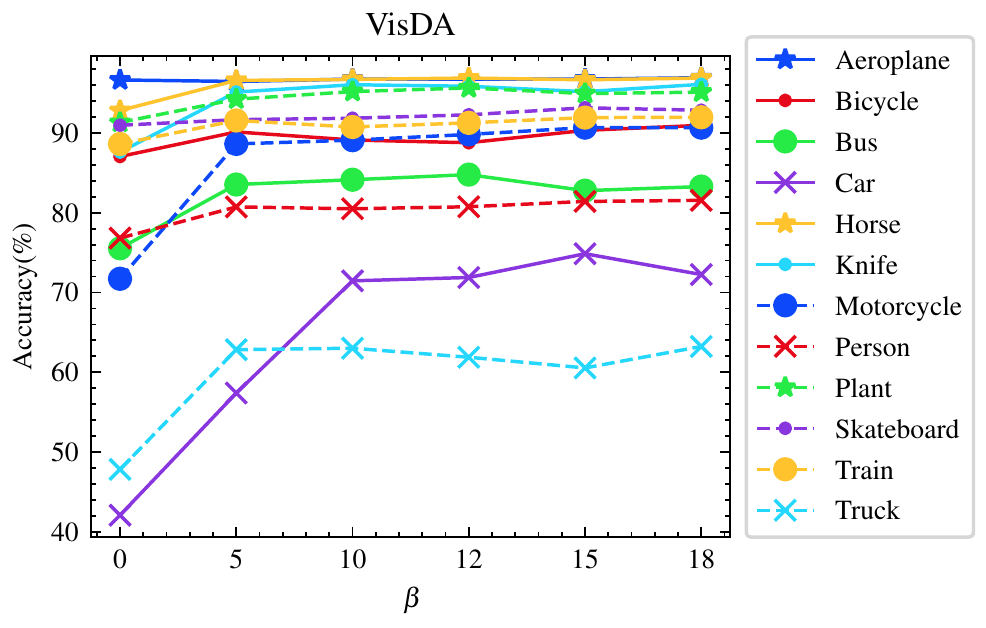}
\caption{Classification accuracy(\%)  on VisDA with different $\beta$ values}
\label{fig:beta-visda}
\end{figure}

During later stages of training, the negative pairs term that pulls apart the samples may dominate the loss values, affecting overall performance. Therefore, we introduce a factor $\alpha = \left( \frac{max\_iter}{max\_iter + iter} \right)^{\beta}$ to regulate the impact of negative pairs. Here, $max\_iter = batch\_size \times epoch$, and $\beta$ controls the rate of decrease. A larger value of $\beta$ results in a faster reduction of the negative pairs impact as the epoch progresses. By introducing $\beta$, the accuracy on VisDA shows steady improvement. Comparison results with and without decay are depicted in Fig.\ref{fig:acc}. To further explore the effect of $\beta$ on VisDA, we display accuracy results for each category at different values in Fig.\ref{fig:beta-visda}. As $\beta$ increases, the accuracy of the two most numerous categories, "car" and "truck", improves significantly, while other categories maintain high accuracy, demonstrating the effectiveness and robustness of the decay factor.

\begin{figure}[htbp]
\centering
\subcaptionbox{Art\label{beta-oh_a}}{
\includegraphics[width = .235\linewidth]{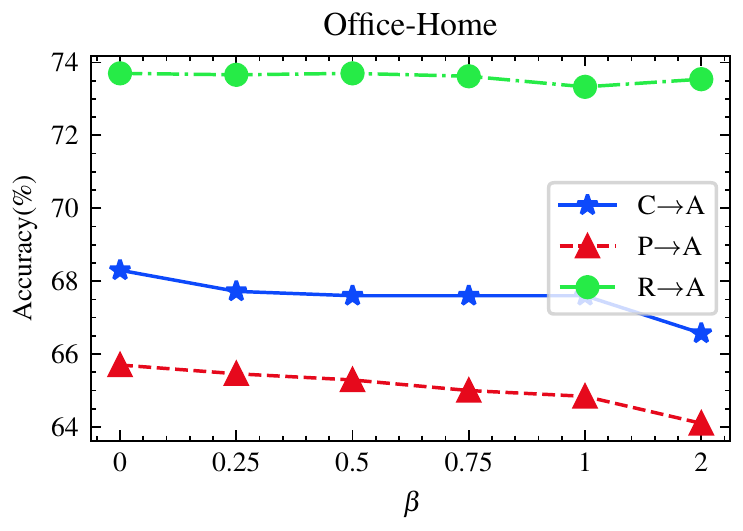}}
\hfill
\subcaptionbox{Product\label{beta-oh_p}}{
\includegraphics[width = .235\linewidth]{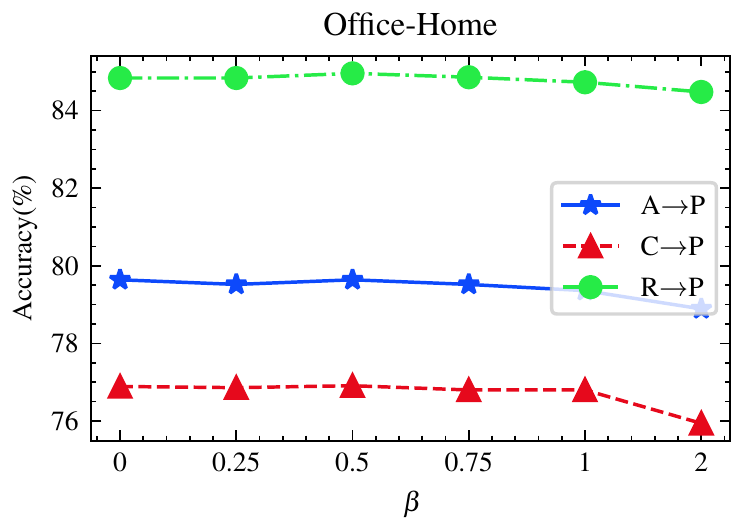}}
\hfill
\subcaptionbox{Clipart\label{beta-oh_c}}{
\includegraphics[width = .235\linewidth]{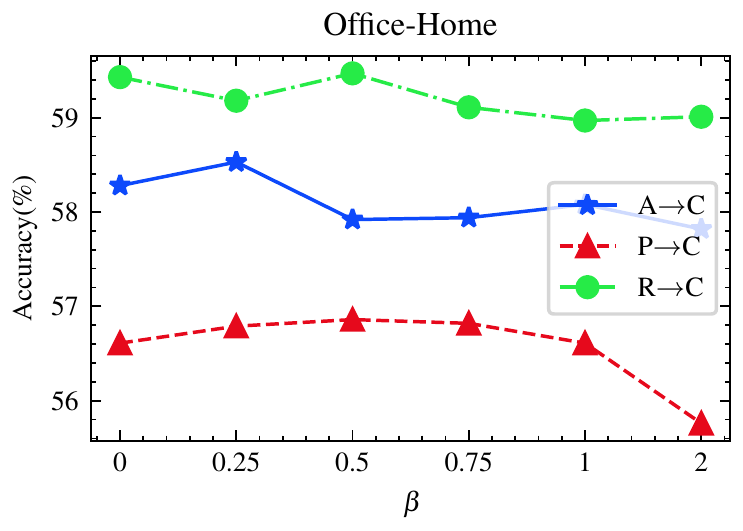}}
\hfill
\subcaptionbox{Real World\label{beta-oh_r}}{
\includegraphics[width = .235\linewidth]{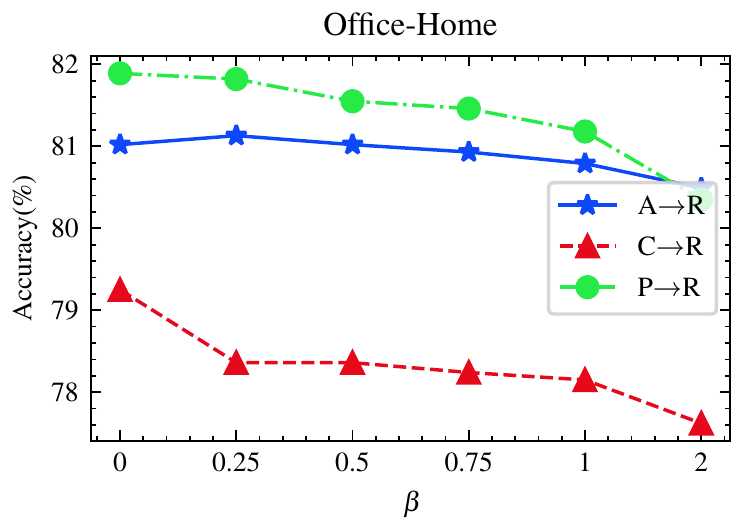}}
\caption{Classification accuracy(\%) on Office-Home with different $\beta$ values}
\label{fig:beta-oh}
\end{figure}

\begin{figure}[htbp]
\centering
\subcaptionbox{Amazon\label{beta-o_a}}{
\includegraphics[width = .31\linewidth]{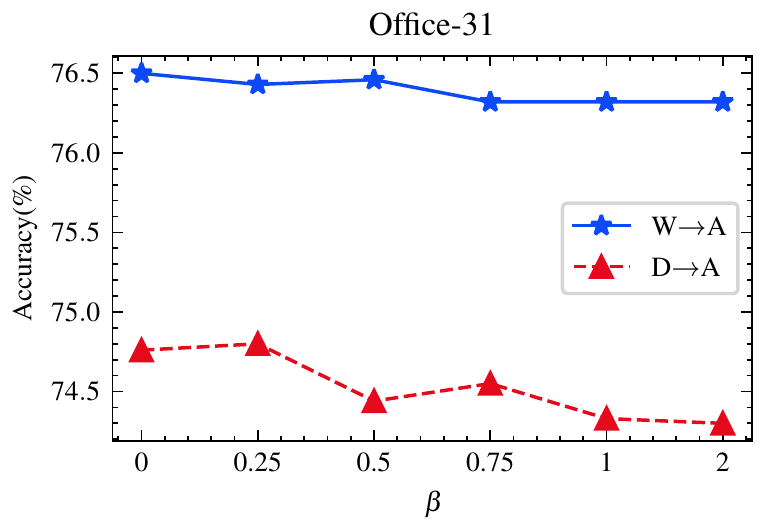}}
\hfill
\subcaptionbox{DSLR \label{beta-o_d}}{
\includegraphics[width = .31\linewidth]{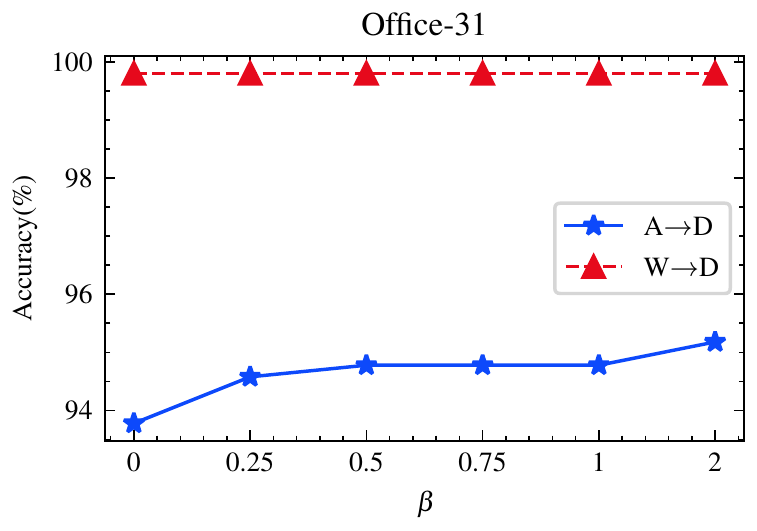}}
\hfill
\subcaptionbox{Webcam \label{beta-o_w}}{
\includegraphics[width = .31\linewidth]{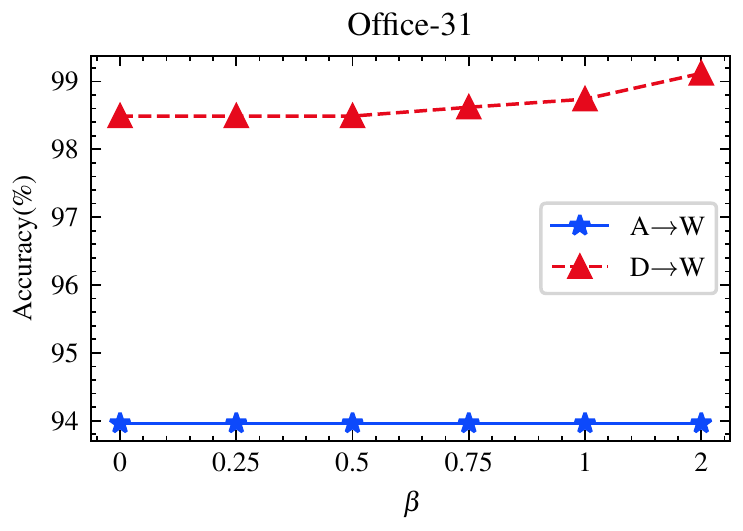}}
\caption{Classification accuracy(\%) on Office-31 with different $\beta$ values}
\label{fig:beta-o}
\end{figure}

Furthermore, we present the results on different values of the decay factor $\beta$ for the other two datasets. For Office-Home, we report the four domains, as shown in Fig.\ref{fig:beta-oh}. And for Office-31 is depicted in Fig.\ref{fig:beta-o}. These two datasets show a robust performance that remains consistently high across various $\beta$ values.

In summary, a key aspect of this improvement revolves around the influence of negative pairs. Initially, these pairs help differentiate dissimilar samples as learning progresses. However, their importance diminishes in the later stages, especially when dealing with imbalanced datasets. This is because excessive separation caused by negative pairs can introduce a harmful bias, leading to a degradation of model performance over time. To address this, the introduction of a decay factor can effectively counteract this bias, resulting in an overall enhancement of the model performance. 

\subsubsection{Number of neighbors K}
\begin{figure}
\centering
\subcaptionbox{Art\label{k-oh_a}}{
\includegraphics[width = .23\textwidth] {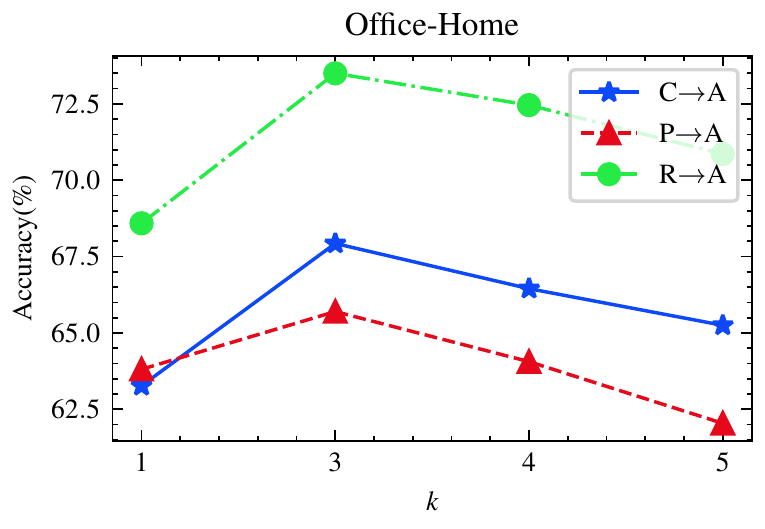}}
\hfill
\subcaptionbox{Product\label{k-oh_p}}{
\includegraphics[width = .23\textwidth] {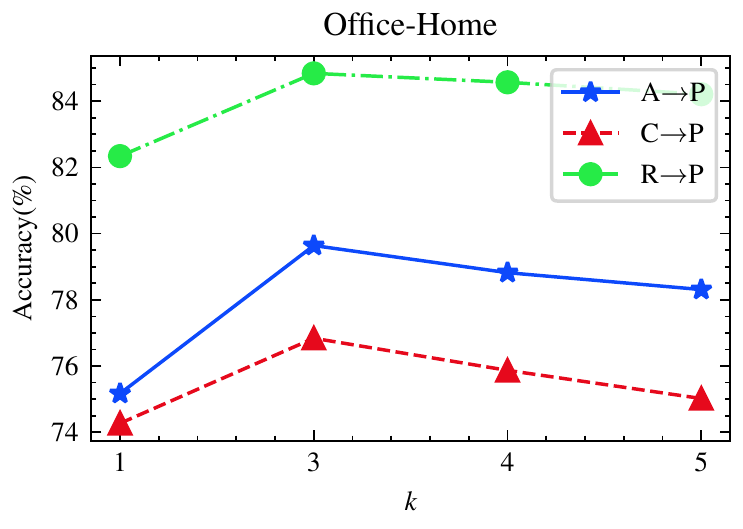}}
\hfill
\subcaptionbox{Clipart\label{k-oh_c}}{
\includegraphics[width = .23\textwidth] {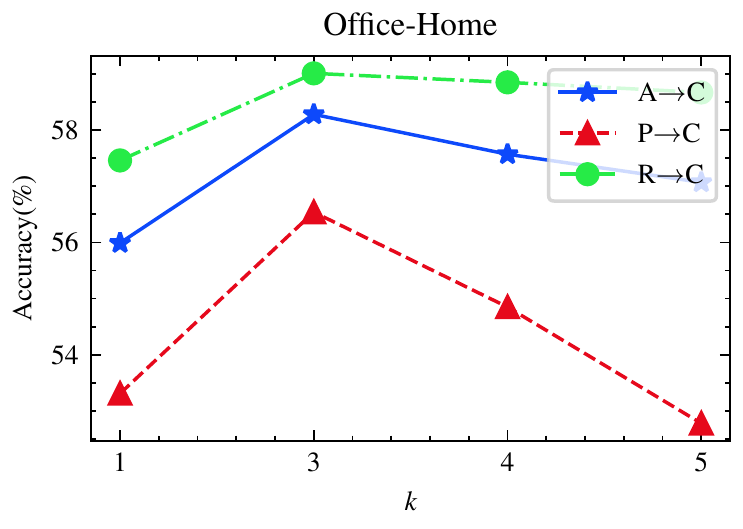}}
\hfill
\subcaptionbox{Real World\label{k-oh_r}}{
\includegraphics[width = .23\textwidth] {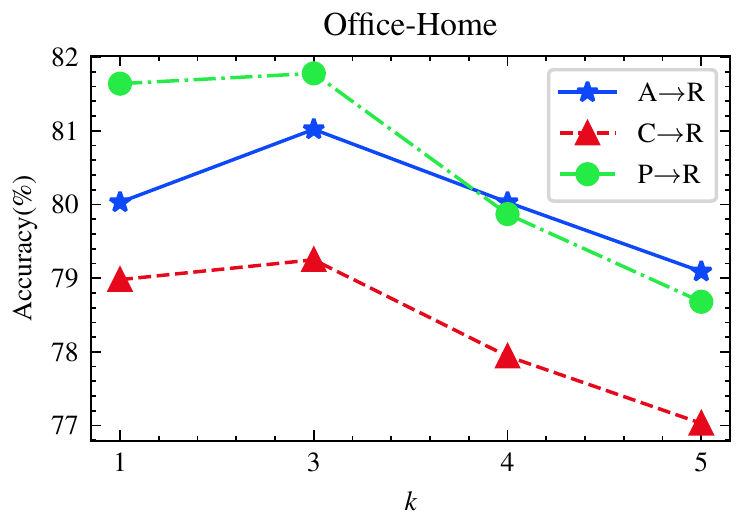}}
\caption{Classification accuracy(\%)  on Office-Home with different K values}
\label{fig:k-oh}
\end{figure}

\begin{figure}
\centering
    \subcaptionbox{Amazon\label{k-o_a}}{
    \includegraphics[width = .31\linewidth] {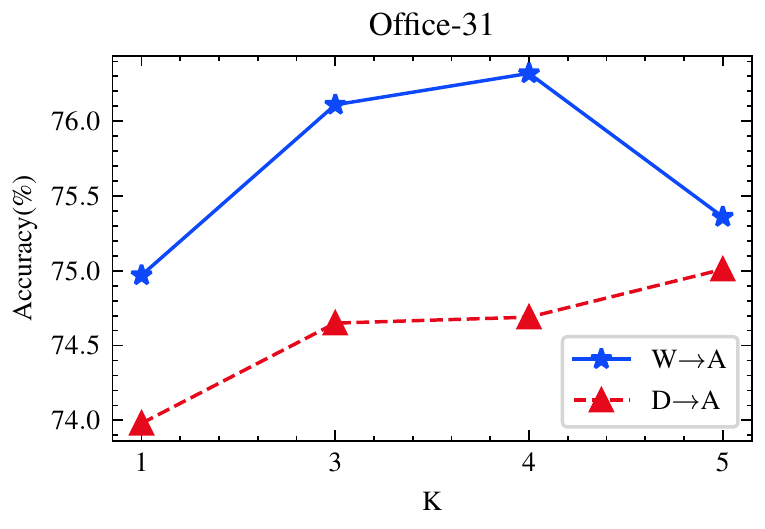}}
    \hfill
    \subcaptionbox{DSLR\label{k-o_d}}{
    \includegraphics[width = .31\linewidth] {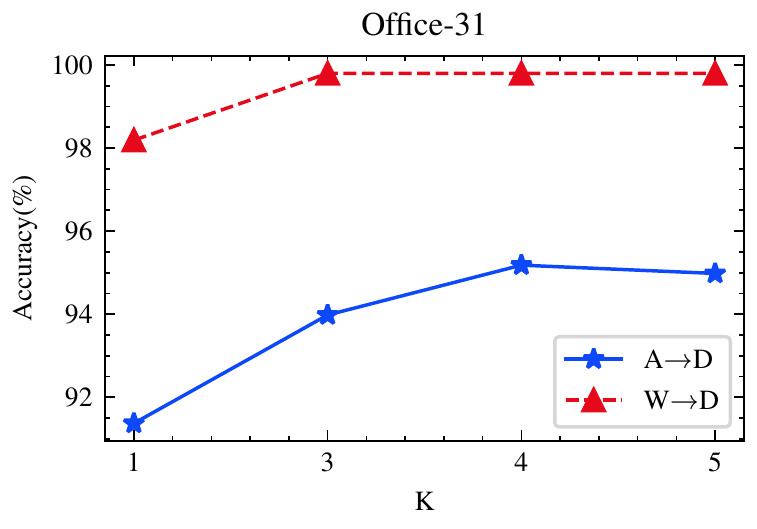}}
    \hfill
    \subcaptionbox{Webcam\label{k-o_w}}{
    \includegraphics[width = .31\linewidth] {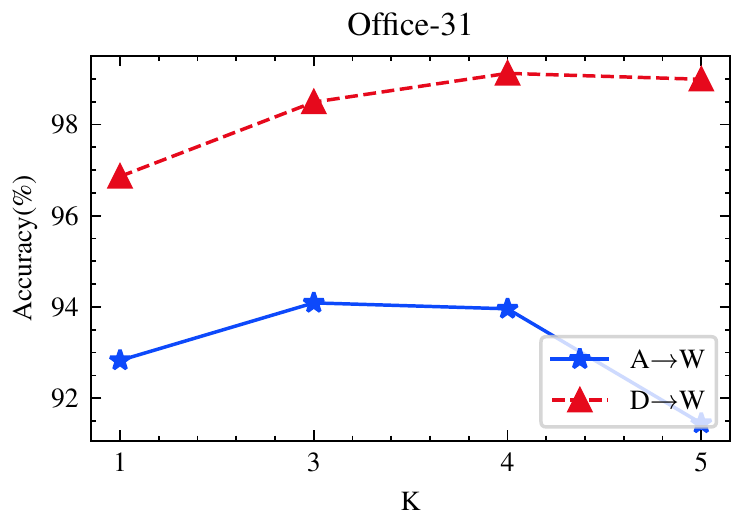}}
    \caption{Classification accuracy(\%)  on Office-31 with different K values.}
    \label{fig:k-o}
\end{figure}

\begin{figure}
\centering
\includegraphics[width = .6\linewidth] {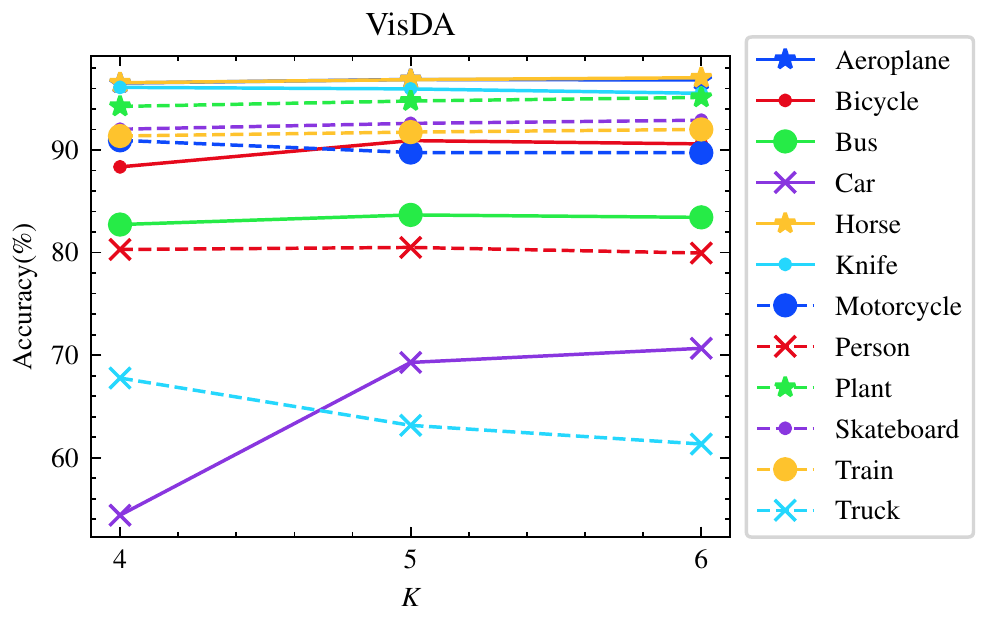}
\caption{Classification accuracy(\%)  on VisDA with different K values.}
\label{fig:k-visda}
\end{figure}

\begin{figure}
\centering
\includegraphics[width = .5\linewidth] {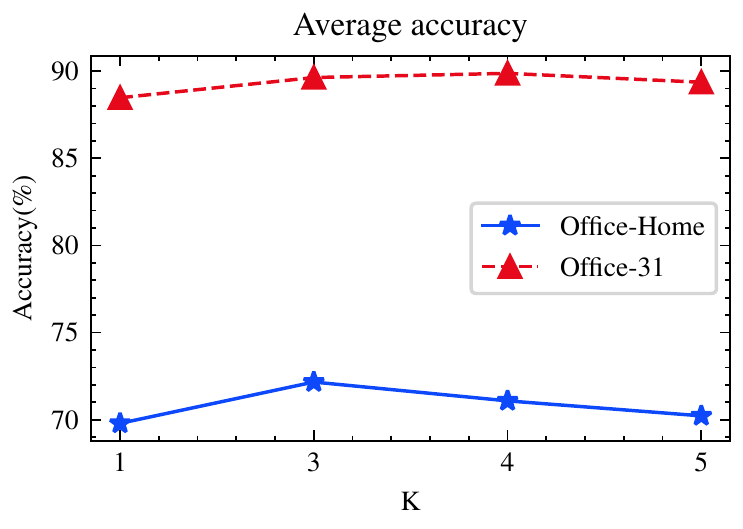}
\caption{Average accuracy(\%)  on Office-Home and Office-31.}
\label{fig:k-avg}
\end{figure}

For the number of neighbors K used for feature clustering in Eq.\eqref{eq_cac}, we show the sensitivity of our method to the parameter K on each of the three datasets in Fig.\ref{fig:k-oh}- Fig.\ref{fig:k-visda}. From Eq.\eqref{eq_cac}, we can see that K is correlated with the sample size of the dataset, requiring a larger value on the larger VisDA dataset and a relatively smaller value for Office-Home. Additionally, we show the average accuracy on Office-Home and Office-31 in Fig.\ref{fig:k-avg}, where the smaller datasets Office-Home and Office-31 obtain higher accuracy when K is 3 or 4. 

For the number of neighbors K used for feature clustering in Eq.\eqref{eq_cac}, we examine the sensitivity of our method to the parameter K on each of the three datasets, as illustrated in Fig.\ref{fig:k-oh} to Fig.\ref{fig:k-visda}. From Eq.\eqref{eq_cac}, it is evident that K is correlated with the sample size of the dataset, necessitating a larger value on the larger VisDA dataset and a relatively smaller value for Office-Home. Additionally, we display the average accuracy on Office-Home and Office-31 in Fig.\ref{fig:k-avg}, where the smaller datasets Office-Home and Office-31 achieve higher accuracy when K is set to 3 or 4.

As observed from the results, larger K values consider more pairs, which is beneficial for learning a robust boundary. However, setting a K value that is too large may also include samples from other categories, introducing more noisy samples and leading to performance degradation.

\subsubsection{Runtime analysis} 

\begin{table}
    \centering
    \caption{Runtime analysis and the average accuracy(\%) of SHOT and our methods on VisDA. 30\%  denote the percentage of target features stored in the memory bank.}
    \label{tab:runtime}
    \begin{tabular}{c|c|c}
    \toprule
    Method & Runtime(s/epoch) & \makebox[0.07\textwidth][c]{\textbf{Avg}}\\
    \hline
    SHOT & 485 & 82.9    \\
    \hline
    CaC(Ours)  & 471 &  \textbf{87.7} \\
    30\% for memory bank & 466  & 87.5 \\
       \bottomrule
       \end{tabular}
\end{table}
We compare the runtime in one epoch with SHOT in Table \ref{tab:runtime}. Regarding SHOT, pseudo-label computation involves clustering in each iteration. In contrast, our nearest neighbor samples are obtained by computing a dot product operation on the samples with the memory bank $\mathcal{F}$, followed by selecting the top-K most similar samples. This process is remarkably simple as the dot product is equal to the cosine distance after feature normalization. Notably, even though weights $W_{sim}$ require the use of extended nearest neighbors, no additional computational consumption is needed because these extended nearest neighbors can be directly retrieved from the neighbor bank $\mathcal{N}$. As a result, our method improves performance with reduced computation. Additionally, we successfully reduce the size of the repository without incurring significant performance loss, ensuring competitive results.
\subsubsection{Feature visualization}
\begin{figure}[htbp]
\centering
\subcaptionbox{Before adaptation\label{fig:oh before}}{
\includegraphics[width = .47\linewidth]{ 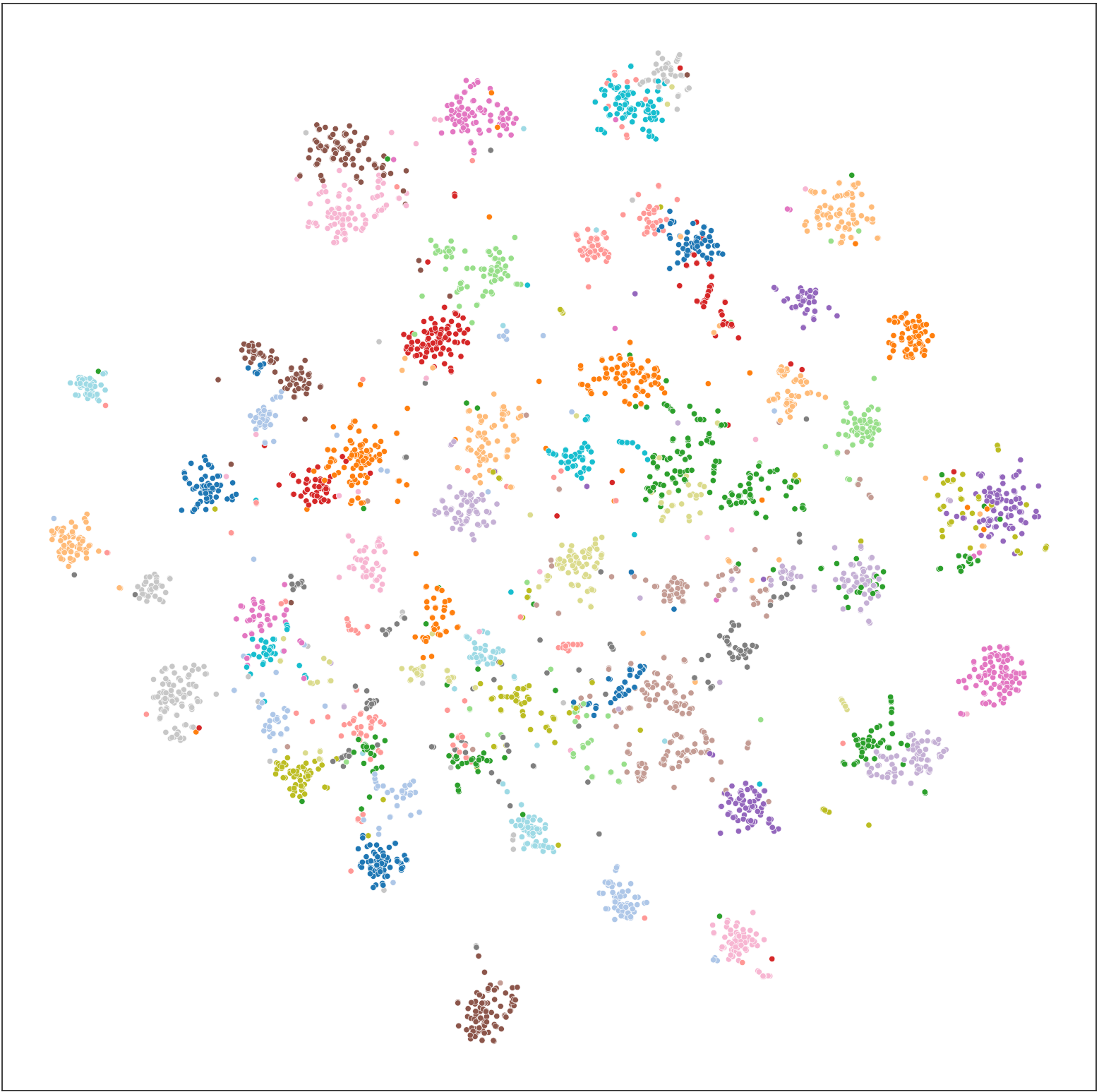}}
\hfill
\subcaptionbox{CaC\label{fig:oh after}}{
\includegraphics[width = .47\linewidth]{ 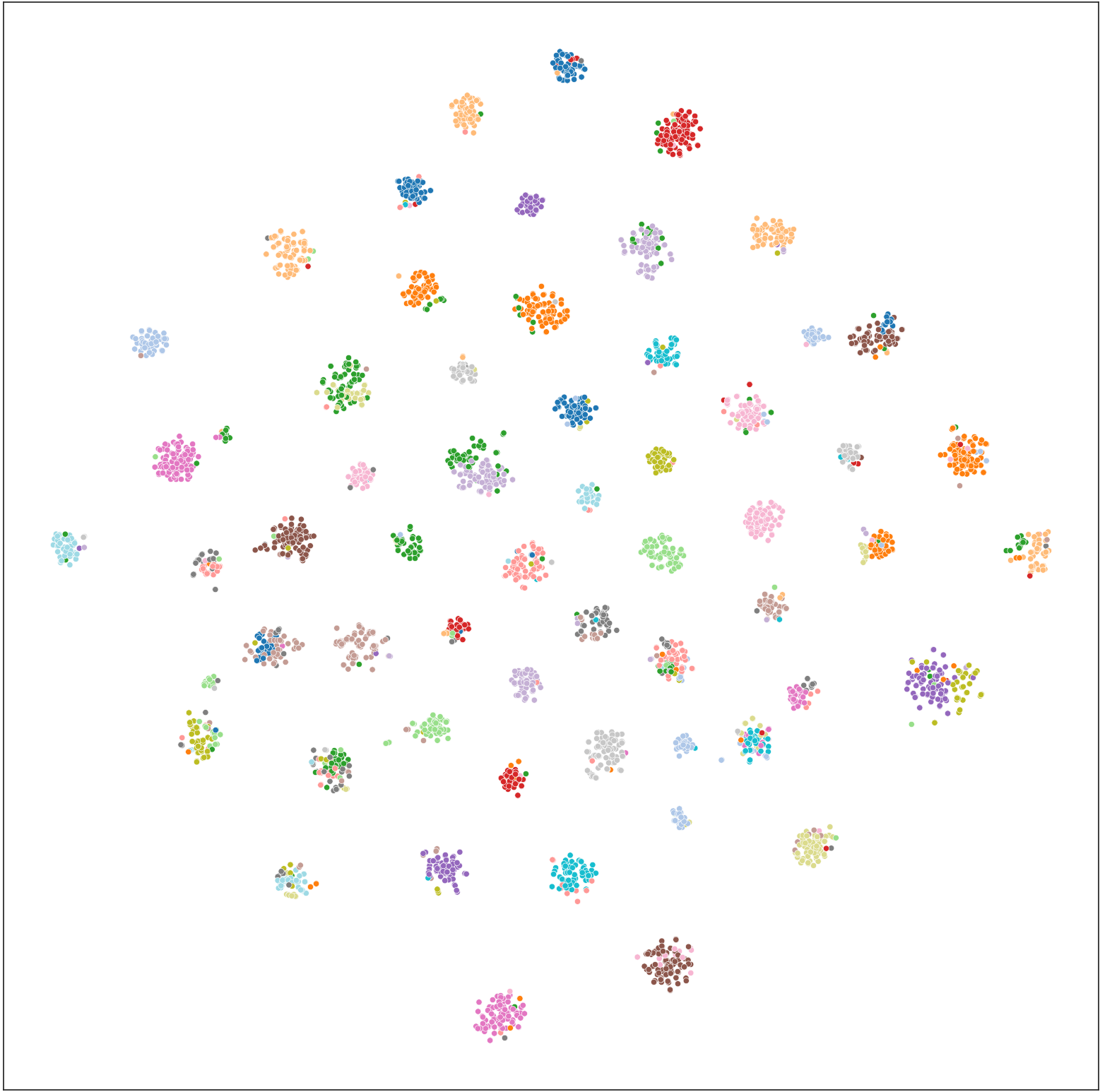}}
\caption{The TSNE visualization on Office-Home.}
\label{fig:tsne oh}
\end{figure}

\begin{figure}[htbp]
\centering
\subcaptionbox{Before adaptation\label{fig:visda before}}{
\includegraphics[width = .47\linewidth]{ 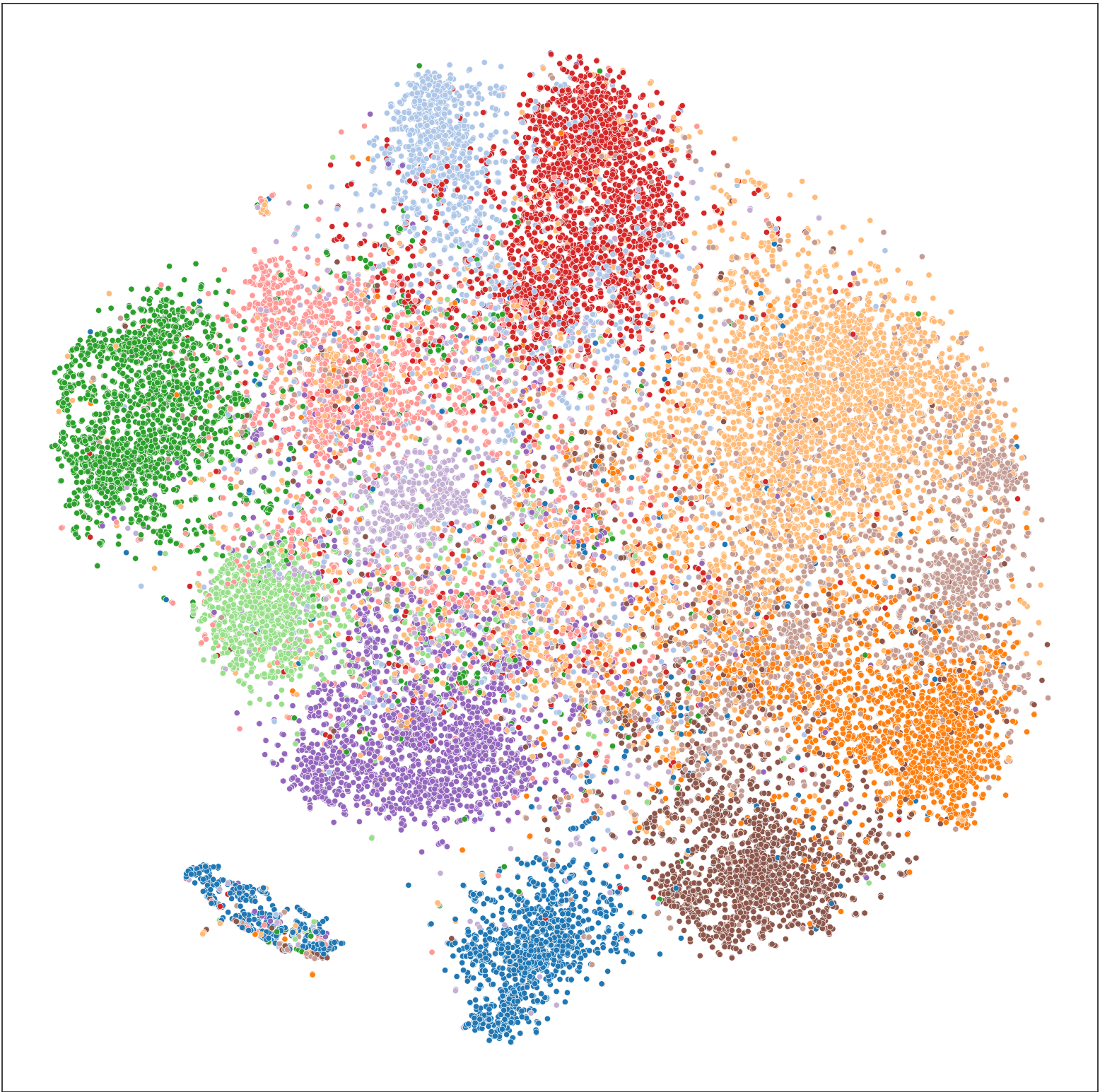}}
\hfill
\subcaptionbox{CaC\label{fig:visda after}}{
\includegraphics[width = .47\linewidth]{ 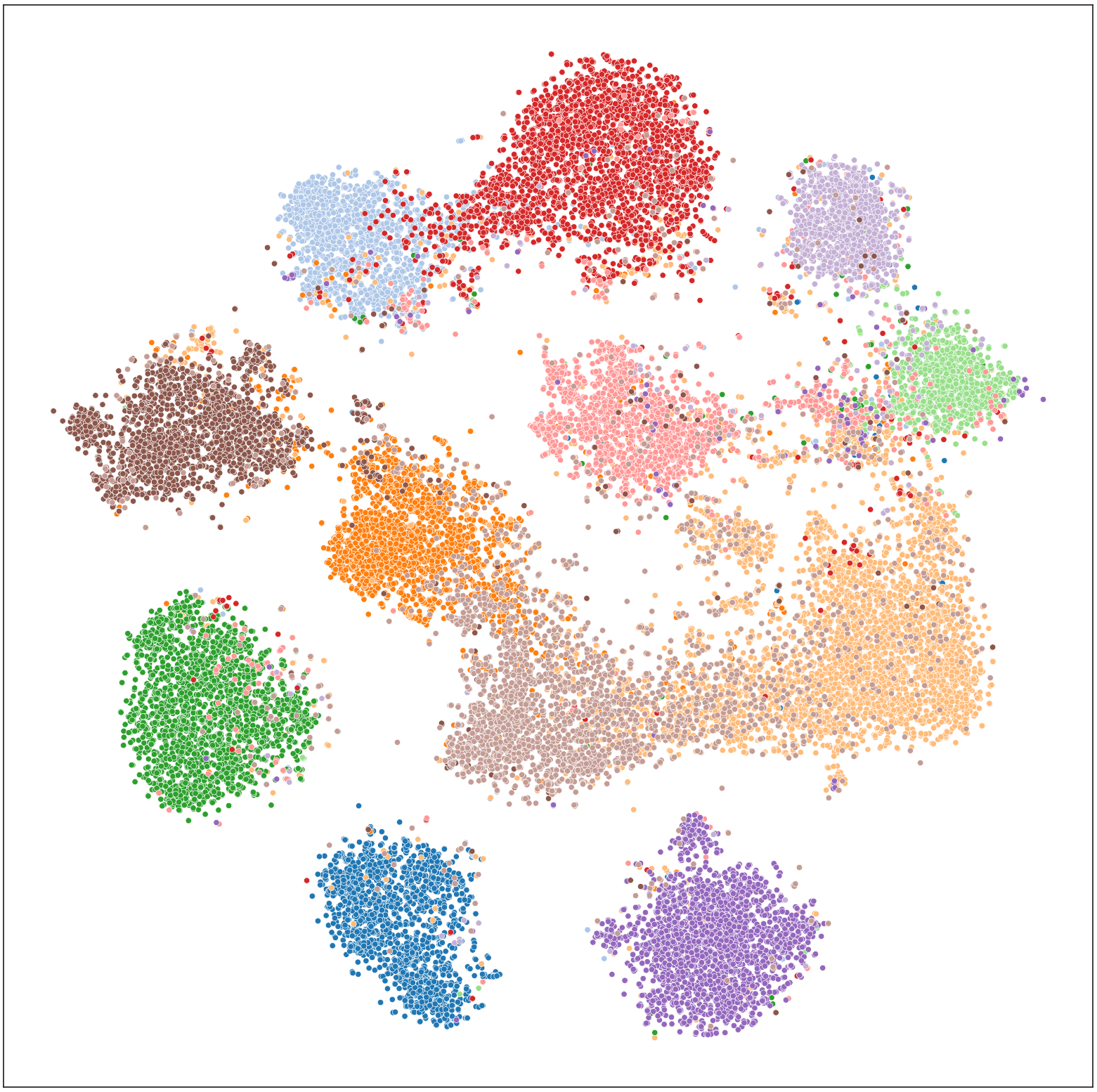}
}
\caption{The TSNE visualization on VisDA.}
\label{fig:tsne visda}
\end{figure}

To further validate the effectiveness of our method, we perform visualization and analysis of the learned features before and after adaptation. The visualization is achieved using the TSNE technique on two datasets: Office-Home and VisDA. For the dataset Office-Home, as shown in Fig.\ref{fig:tsne oh}, which has a larger number of classes, CaC demonstrates improved intra-class compactness and inter-class separation. Likewise, on the more challenging synthetic-to-real dataset VisDA, as illustrated in Fig.\ref{fig:tsne visda}, the classification boundaries become more apparent due to our contrastive learning strategy.

\section{Conclusion}\label{sec13}
In this paper, we present an innovative technique for unsupervised domain adaptation that does not require access to source data. Our method is designed to acquire domain-invariant features by emphasizing the consistency of outputs with nearest neighbors. We establish an indexed memory bank to store nearest neighbor samples, enabling efficient retrieval of extended neighbors. These extended neighbors are then employed to identify more valuable negative pairs, without adding to the computational load. Rigorous experiments confirm the substantial impact of our proposed approach on both positive and negative pairs. Furthermore, the analysis of the negative term's influence holds promise for enriching future works in contrastive learning. These findings ultimately showcase its superior performance when compared to other source-free domain adaptation methods, as demonstrated across three benchmark datasets.

\section*{CRediT authorship contribution statement}
\textbf{Yuqi Chen:} Methodology, Data curation, Visualization, Investigation, Writing-original draft. \textbf{Xiangbin Zhu:} Supervision, Validation. \textbf{Yonggang Li:} Supervision, Writing-review\&editing. \textbf{Yingjian Li:} Validation. \textbf{Haojie Fang:} Validation.
\section*{Declaration of competing interest}
The authors declare that they have no known competing financial interests or personal relationships that could have appeared to influence the work reported in this paper.
\section*{Acknowledgement}
This work was partially supported by the National Natural Science Foundation of China (NSFC Grant No. 61972059), Provincial Natural Science Foundation of Zhejiang (Grant No. LY19F020017), Jiaxing Science and Technology Project (Grant No.2023AY11047).
\bibliographystyle{elsarticle-num}
\bibliography{cas-refs}







\end{document}